\documentclass[opre]{informs4}
\RequirePackage{tgtermes}
\RequirePackage{newtxtext}
\RequirePackage{newtxmath}
\RequirePackage{bm}
\RequirePackage{endnotes}
\usepackage{tabularx}
\usepackage[export]{adjustbox}
\usepackage{caption}
\usepackage{makecell}
\usepackage{booktabs}

\renewcommand{\qed}{\hfill\blacksquare}

\OneAndAHalfSpacedXII 

\usepackage{tikz}

\usepackage{algorithm}
\usepackage{algorithmic}

\usepackage{natbib}
 \bibpunct[, ]{(}{)}{,}{a}{}{,}%
 \def\bibfont{\small}%

\newcommand{\paren}[1]{\left( #1 \right)}
\newcommand{\R}[0]{\mathbb{R}}

\newcommand{\norm}[1]{\left\lVert#1\right\rVert}

\newcommand{\x}[0]{\mathbf{x}}
\newcommand{\y}[0]{\mathbf{y}}
\renewcommand{\v}[0]{\mathbf{v}}
\newcommand{\z}[0]{\mathbf{z}}
\newcommand{\Z}[0]{\mathbf{Z}}

\newcommand{\newpara}[1]{\vskip 0.25cm \noindent \textbf{#1}}

\def\phcomments{1}
\newcounter{note}[section]
\renewcommand{\thenote}{\thesection.\arabic{note}}
\newcommand{\note}[2]{\refstepcounter{note}\marginpar{\small\bf \textcolor{red}{#1~\thenote}}$\ll${\sf \textcolor{red}{#1's
 Comment~\thenote:}} {\em \textcolor{red}{#2}}$\gg$}

\ifnum\phcomments=1
\newcommand{\notePH}[1]{\note{PH}{#1}}
\else
\newcommand{\notePH}[1]{}
\fi
\usepackage{soul}

\EquationsNumberedThrough    

\TheoremsNumberedThrough     
\ECRepeatTheorems  %

\MANUSCRIPTNO{MOOR-0001-2024.00}

\begin{document}


\RUNAUTHOR{Cristian et al.}

\RUNTITLE{Aligning Learning and Endogenous Decision-Making}

\TITLE{Aligning Learning and Endogenous Decision-Making}

\ARTICLEAUTHORS{%
\AUTHOR{Rares Cristian}
\AFF{Massachusetts Institute of Technology, \EMAIL{raresc@mit.edu}} 
\AUTHOR{Pavithra Harsha}
\AFF{IBM Research \EMAIL{pharsha@us.ibm.com}}
\AUTHOR{Georgia Perakis}
\AFF{Massachusetts Institute of Technology \EMAIL{georgiap@mit.edu}}
\AUTHOR{Brian Quanz}
\AFF{IBM Research \EMAIL{blquanz@us.ibm.com}}
} 

\ABSTRACT{%
Many of the observations we make are biased by our decisions. For instance, the demand of items is impacted by the prices set, and online checkout choices are influenced by the assortments presented. The challenge in decision-making under this setting is the lack of counterfactual information, and the need to learn it instead. We introduce an \emph{end-to-end} method under endogenous uncertainty to train ML models to be aware of their downstream, enabling their effective use in the decision-making stage. We further introduce a robust optimization variant that accounts for uncertainty in ML models --- specifically by constructing uncertainty sets over the space of ML models and optimizing actions to protect against worst-case predictions. We prove guarantees that this robust approach can capture near-optimal decisions with high probability as a function of data. Besides this, we also introduce a new class of two-stage stochastic optimization problems to the end-to-end learning framework that can now be addressed through our framework. Here, the first stage is an information-gathering problem to decide which random variable to ``poll'' and gain information about before making a second-stage decision based off of it. We present several computational experiments  for pricing and inventory assortment/recommendation problems.  We compare against existing methods in online learning/bandits/offline reinforcement learning and show our approach has consistent improved performance over these. Just as in the endogenous setting, the model's prediction also depends on the first-stage decision made. While this decision does not affect the random variable in this setting, it does affect the correct point forecast that should be made.
}%





\maketitle


\section{Introduction}\label{sec:Intro}

The alignment of AI systems with real-world objectives is critical for making robust and efficient systems, especially as we continue to integrate AI into increasingly complex decision-making. In practice, machine learning models serve as first-stage forecasters of unknown quantities, and their predictions feed into a second-stage optimization problem to drive the final decisions. 
In this paper, we will focus on decision problems with \emph{endogenous} uncertainty, settings in which outcomes depend directly on the decision themselves, with the goal of aligning learning and decision-making. 
To illustrate this workflow, consider the following prototypical pricing problem, where in the product demand is influenced by product price: (1) we learn a model to predict demand as a function of price; (2) we define the objective value (revenue) as a function of both the decision (price) and the predicted random variable (demand); and (3) we optimize over all possible price decisions and choose the one with maximum revenue based on our demand predictions. In this paper, we consider how to learn demand (the endogenous random variable), and importantly how to make this learning process aware of its impact on the downstream optimization step. 

\begin{figure}[b]
    \centering
    \includegraphics[width=1\linewidth]{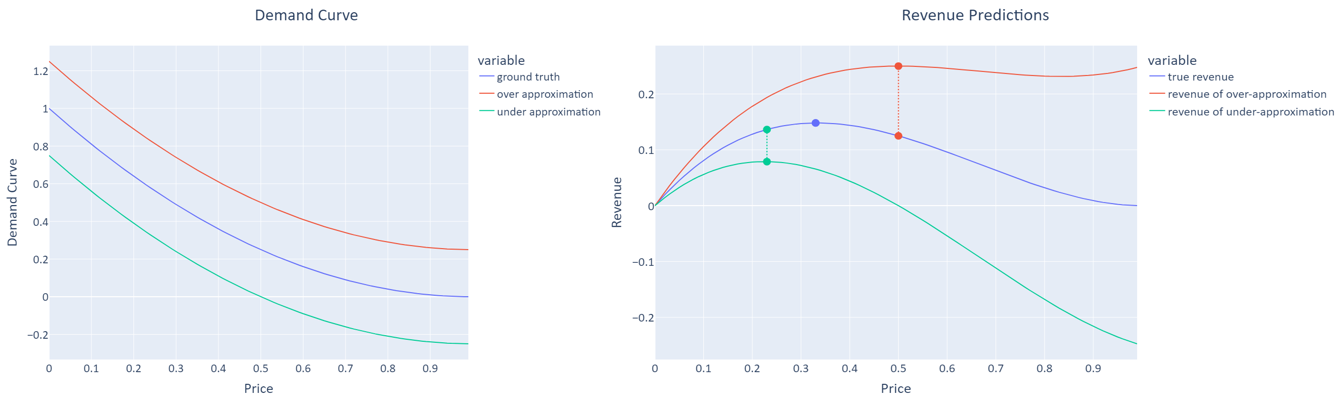}
    \caption{(left) two demand predictions as a function of price that have the same mean-squared error compared to true demand, the blue curve; (right) revenue prediction corresponding to each demand prediction.}
    \label{fig:pricing-ex}
\end{figure}

A traditional approach would separate the learning and decision making steps. However, this brings several issues. For instance, when optimizing over the entire input space, it becomes easy to choose decisions that are far out of sample and for which the model has poor predictive power. This may result in decisions with objective significantly worse than predicted. As an example for pricing, the actual demand could be significantly lower than predicted. This is especially problematic when there is sparse or limited data. Consider the following example in Figure \ref{fig:pricing-ex}. On the left, we have two predictions of demand as a function of price -- both are equally good in the sense that each has the same mean-squared error compared to true demand. On the right, we have the corresponding prediction of revenue (revenue being price multiplied by corresponding predicted price). However, while demand prediction was equally good, the corresponding revenue predictions are not. The decision taken by maximizing revenue as predicted by the green curve is better than the one taken from using the red curve. The end-to-end approach we propose would be able to identify in the training stage which demand prediction model performs better on the final task-based objective.

In the setting of exogenous uncertainty, the outcomes are independent of our decision. For example, online demand for a product is independent of inventory allocation decisions in a warehouse, even though demand predictions are crucial for inventory decisions. A common approach to aligning the predictions with the optimization stage is end-to-end learning, where the two stages are integrated into a single trainable pipeline. The objective function of the learning problem is to directly maximize the value of the corresponding decisions taken. However, existing approaches cannot be applied to the endogenous uncertainty setting as the end-to-end formulation requires knowledge of the outcome or value of a decision taken. When the uncertainty is independent of the decision taken, one can simply use the data as the ground truth. However, when the uncertainty depends on the decision, one can no longer take this approach since one does not have access to counterfactual information. That is, we do not have knowledge about what would have happened if a different action were taken. 

We introduce a cost-aware end-to-end method for this initial learning stage in the endogenous uncertainty case. Specifically, the objective is to train a model whose corresponding predictions on in-sample decisions have task-based cost close to the observed cost in-sample. This allows the model to learn, for example, the relevant statistics of the known distribution that are relevant for the specific decision task.

Next, we build a robust method on top of this. There are many models that can be learned through the end-to-end cost-aware method above, and our robust approach considers them all simultaneously. Intuitively, given some set of data there are many different predictive functions or models that are \emph{consistent} with the data in the sense that they have low in-sample error. However, these consistent models might behave differently on unseen or counterfactual situations (i.e., outcomes we didn’t observe). For each data-point we propose to find, among all such \emph{consistent} predictors, a function that produces the lowest predicted objective value for that data-point. Then, we choose the decision which maximizes this lower bound. Unlike traditional approaches, where we fix a predictor here were are instead working with a class of predictors and pick a predictor-action pair that maximizes the worst case reward. 
We provide a full summary of  contributions around theoretical results and algorithms for both approaches in section \ref{sec:contributions}. 

\subsection{Information Gathering}

We introduce a new problem domain under the endogenous end-to-end paradigm that to our knowledge has not been considered before in the literature. This new setting, which we will refer to as the information-gathering problem, consists of
an initial phase to gather information about some of the random variables before the prediction and decision-making steps. For example, a survey can be sent or a poll can be set up to better understand demand at a particular location. Given this new information, one can then make more informed predictions about the rest of the random variables (for e.g. demand at the remaining locations) and subsequently make a more informed decision. If demand across locations is correlated, one can gain significant information by polling a single location. As a first stage, one must decide which location to poll, observe information about this location, then make a new prediction and a decision for all locations conditioned on this new observation. So, there are three questions (1) which random variable should we poll, (2) what predictions to make conditioned on observing this chosen variable, and (3) what decision to make based off of these predictions.

In contrast to the first endogenous uncertainty setting we presented, the random variables here are not dependent on the decisions we take. However, our knowledge of the random variable does depend on the first stage polling decision we take. 
We introduce a novel approach to end-to-end learning for this two-stage information-gathering problem by unifying our endogenous end-to-end framework with the exogenous end-to-end methods. 

\subsection{Contributions}
\label{sec:contributions}

\begin{enumerate}
    \item We formulate an end-to-end learning approach for the endogenous uncertainty case, when the realizations of uncertainty are dependent on the decisions taken. The objective is to train a model whose corresponding predictions on in-sample decisions have task-based reward close to the observed reward in-sample. In addition, due to the non-convex nature of the end-to-end objective, we provide mixed-integer optimization formulations, as well as a computationally efficient sampling-based approach. See section \ref{sec:algorithms1}. 
    \item We provide theoretical analysis showing that (1) optimally minimizing our proposed objective produces optimal decisions and (2) convergence bounds on the generalization gap with respect to the amount of training data used and model complexity. Overall, the generalization gap between in-sample and out-of-sample cost decreases as $1/\sqrt{N}$ where $N$ is the amount of training data. 
    \item We propose a robust optimization extension to the end-to-end approach under endogenous uncertainty. We construct an uncertainty set of possible models that ``fit the data'' well. We take decisions by maximizing the worst-case function from the uncertainty set. We prove that this choice of uncertainty set contains the true optimal predictive function with high probability, and the size of the set can be decreased linearly as a function of the amount of training data available.
    \item We propose an algorithm for the robust optimization formulation based on cutting-planes to solve the outer maximization problem. We also propose an algorithm for the inner problem, finding the worst case function over the uncertainty set for a fixed action. This is based on alternatingly minimizing the objective and projecting back into the feasible uncertainty set.
    \item We extend our proposed end-to-end method to the information-gathering problem. This combines traditional end-to-end methods under exogenous variables in the second-stage problem and under endogenous variables in the first stage problem. This two-stage problem class has not been studied under the end-to-end learning setting to date.  
    \item We show computational experiments on pricing and assortment optimization problems. We  assume the demand of a single product is dependent on decisions taken for all other items due to complementary or supplementary effects. We show the end-to-end approach, as well as the robust counterpart,
    improves significantly over a traditional two-stage approach and other robust offline learning methods. Finally, we also consider an electricity generation and scheduling problem under information-gathering. Here we make an initial forecast, and must decide on the optimal time to update the forecast. The amount of time we decide to wait will provide us with different gathered information that we can use to make forecasts for the remaining schedule. 
    This involves learning how to balance the benefits of waiting for more accurate information against the costs of delaying decisions.
\end{enumerate}

\subsection{Related Work}

The traditional approach is a two step, or two stage, method. That is, first independently learn a model of the random variable as a function of features and decisions, then afterwards optimize to make a decision. This has been applied to a wide variety of problem areas, like \cite{ettl2020data, kash2023optimal} in personalized pricing and recommendation, \cite{cohen2019optimizing} in promotion optimization, and assortment optimization \cite{akchen2021assortment, agrawal2019mnl}.


The closest connection to our endogenous end-to-end problem is the exogenous case. Here, the decisions do not affect the observation. This setting allows for more flexibility since we can take any decision, and are able to evaluate its cost using the training data. In the endogenous setting, this cannot be done since we need counterfactual information that is unavailable. The exogenous case has had considerable study, perhaps starting in \cite{elmachtoub2022smart} for linear problem, with extensions to a variety of settings: \cite{vlastelica2019differentiation}, \cite{berthet2020learning} for other approaches to linear problems, \cite{amos2017optnet} for quadratic problems, \cite{agrawal2019differentiable} for general convex problems, \cite{cristian2023end} for more time-efficient meta-optimization methods, \cite{ferber2020mipaal}, \cite{mandi2020smart} for hard/integer problems. See for instance \cite{sadana2023survey} for a more comprehensive survey.

The space of online learning also has striking similarities to the offline case with endogenous uncertainty. Indeed, a major aspect of the contextual online problem is to learn the rewards associated with actions by sequentially choosing actions to minimize regret or enhance the speed of learning. Linear UCB \citep{chu2011contextual} for example aims to learn a linear function of the actions and features to predict the associated mean reward. One can view the UCB approach as follows: constructing a confidence bound for each decision and selecting the action that maximizes the upper bound of these estimates, thereby balancing exploration and exploitation. As we select the same action more, our confidence of its value improves, and hence the confidence bound gives some advantage to less-chosen actions to implicitly promote exploration.
Generalizing to a broader context, \cite{foster2021statistical,foster2020beyond} introduce the estimation-to-decision-making meta-algorithm. This approach does not explicitly construct uncertainty sets, however this can be thought of as a learner trying to optimally balance between the regret of their decision with the amount of information they gain (or uncertainty they reduce) from taking a potentially sub-optimal decision. In contrast, our approach does not focus on learning the reward, but rather the intermediate uncertainty. This helps not only to reduce learning complexity, but also to improve interpretability. 

Offline reinforcement learning is also closely related to learning under endogenous uncertainty. For example, while learning the reward of an action, or its Q-value, a model may make errors on out-of-distribution actions and erroneously predict high values. In online reinforcement learning, one would correct this mistake by taking such an action and observing the true value of its reward. However in the offline setting, one is unable to interact with the environment. To combat this issue, some methods aim to constrain the learned policy away from such out-of-distribution actions (for example \citealt{kumar2019stabilizing, fujimoto2019off}). Robustness can also be helpful in the online case \cite{kim2024doubly}. Others more similar to our approach we propose in this paper aim to learn a Q-function that is a lower bound to its true value everywhere (and then choose the action with highest lower bound). See for instance \cite{kumar2020conservative, xie2021bellman}. Our proposed approach differs in that we do not aim to learn any reward function, but rather the underlying uncertain parameter. We also focus more on the algorithmic aspects of solving the more structured problem we consider. 

Versions of the information-gathering problem have been studied as well. The most closely related line of work is in the area of value of information by~\cite{howard1966information}. This aims to decide the amount a decision maker would be willing to pay for information prior to making a decision. This notion of value information is particularly relevant in reinforcement learning applications, deciding which actions to explore to gain the most useful information (for e.g., \citealt{arumugam2021value}). This work still differs significantly as ours also considers the specific structure of the objective and intermediate random variables as we explained previously.

\section{Endogenous end-to-end learning}
\label{sec:framework}

We first formally describe the problem. One wishes to makes decisions $\v \in \mathcal{V}$ in feasible region $\mathcal{V} \subset \R^d$. Associated with this decision is a cost function $c(\v, \z)$ dependent on a random variable $\z$. In turn, the distribution of $\z$ also depends on the decision $\v$ itself and additional contextual information $\x$. We say $\z$ is distributed according to some unknown distribution $\Z|\v,\x$. 


We are given offline data consisting of information $(\x_1, v_1, \z_1), \dots, (\x_n, \v_n, \z_n)$ with decisions $\v_i$ (which are potentially suboptimal) and observed uncertainty $\z_i \sim \Z|\v_n, \x_n$. We make the following assumption on how the data itself is generated.
\begin{assumption}[Data generation]
\label{ass:data}
    We assume the data $(\x_1, \v_1, \z_1), \dots, (\x_N, \v_N, \z_N)$ is generated according to the following process. The feature data $\x_1, \dots, \x_N$ are independent identically distributed random variables from an unknown fixed distribution. Each decision $\v_i$ is generated by an algorithm that depends only on the observations of its history $\mathcal{H}^{n-1}=\{ (\x_1, \v_1, \z_1), \dots (\x_{n-1}, \v_{n-1}, \z_{n-1}) \} $. Finally, $\z_n$ is a sample from the distribution $Z|\v_n,\x_n$ depending on both the decision taken and the features observed. 
\end{assumption}
Notice that this is how decisions are taken in an online setting --- one takes a decision based on past decisions and corresponding observations. However in the offline setting setting we consider we have no control over how the historical decisions are chosen. 

Finally, our objective is to learn some function $f(\x, \v)$ that predicts the uncertainty $\z$ in such a way that the corresponding decisions have minimum cost. In particular, given $f$ one takes decision given by finding $\v$ that minimizes the cost:
\begin{equation}
\label{offline-dec}
    \v^*(f; \x) = \arg \min_{\v \in \mathcal{V}} c(\v, f(\x, \v)).
\end{equation}
Then we wish that the objective of $\v^*(f; \x)$ compared to the optimal decision is minimized:
$\mathbb{E}_{\z \sim \Z|\x, v^*(f; \x)} [c(v^*(f; \x), \z)] - \min_{\v \in \mathcal{V}} \mathbb{E}_{\z \sim \Z|\v,\x} [c(\v, \z)]$. Alternatively, one can view the problem as follow. One is interested in learning some relationship between decisions $\v$ and the uncertainty $\z$. Ideally, we would have access to some function $f^*(\x, \v)$ for which 
\begin{equation}
\label{eq:expectation}
    c(\v, f^*(\x, \v)) = \mathbb{E}_{\z \sim \Z|\x, \v} [c(\v,\z)].
\end{equation}
Then, solving \eqref{offline-dec} using $f^*$ would exactly find the optimal solution. However, one of course does not know $f^*$ and must learn it from data. To achieve this, one can solve the mean-squared error minimization problem below:
\begin{equation}
\label{eq:end-to-end}
    \min_{f \in \mathcal{F}} \sum_{n=1}^N \paren{c(\v_n, f(\x_n, \v_n)) - c(\v_n, \z_n)}^2
\end{equation}
where we must choose a model class $\mathcal{F}$. In general, we will be choosing a class of neural networks and learning the weights. Note that there may not be any $f^* \in \mathcal{F}$ for which $c(\v, f^*(\x, \v)) = \mathbb{E}_{\z \sim \Z|\x, \v}[c(\v, \z)]$ as we wanted in \eqref{eq:expectation}. But as neural networks are approximate universal function learners, as we have enough data we can get close to such an optimal $f^*$. Proposition \ref{prop:exist} proves that such a forecast does exist and Theorem \ref{thm:generalization} describes how much data we may need to learn it.

\noindent For ease of notation going forward, we denote this objective function as the task-based loss:
\begin{definition}[Task-loss]
\label{def:task-loss}
    Let $\mathcal{E}(f)$ denote the task loss of model $f$ on the data $(\x_1, \v_1, \z_1), $ $\dots$, $(\x_N, \v_N, \z_N)$. We define this as:
    \begin{equation}
        \mathcal{E}(f) =  \sum_{n=1}^N \paren{c(\v_n, f(\x_n, \v_n)) - c(\v_n, \z_n)}^2.
    \end{equation}
\end{definition}
We denote the problem in \eqref{eq:end-to-end} of finding an $f$ by minimizing the above task-based loss as an \emph{end-to-end} approach.  We remark that it is crucial to learn an $\hat{f}$ with this task-based loss. To illustrate this, let us examine the alternative approaches below, which will also serve as baselines in our experimental evaluation in section~\ref{sec:experiments}. 

\newpara{Two-stage methods.} The common approach in ML would be to learn a model which learns the mean of the distribution of $\z$. That is, minimize the mean-squared error between predictions and historical observations. We will denote this as a two-stage approach, first predicting $\z$ independent of the task loss $c$, then optimizing for the optimal decision. The two-stage predictor,
$\hat{f}_{\text{2-stage}}$, is a predictor of the mean $ \mathbb{E}[\z] $. The issue arises in the second stage when optimizing.  In general $\mathbb{E}[c(\v,\z)] \not= c(\v,\mathbb{E}[\z])$ when $c$ is non-linear in $\z$.  Therefore, optimizing $\max_{\v\in\mathcal{V}} c(\v, \hat{f}_{\text{2-stage}}(\x, \v))$ would be a proxy for optimizing $\max_{\v\in\mathcal{V}} c(\v,\mathbb{E}_{\z \sim \Z|\v,\x}[\z])$ but not the true objective which is $\max_{\v\in\mathcal{V}} \mathbb{E}_{\z\sim Z|\v,\x} [c(\v,\z)]$.

\newpara{Learning reward directly.} Many methods in online learning such as contextual bandits or reinforcement learning learn the reward function directly, instead of the intermediate random variable $\z$. This removes the issues in the previous section about learning the mean of $\z$ since here we directly optimize the reward.  However, there are several computational downsides to reward-learning. Here, we learn a mapping $r(\v,\x) \approx \mathbb{E}_{\z\sim \Z|\v,\x}[c(\v,\z)]$ while only observing $\v^n, \x^n$ and $c(\v^n,\z^n)$ and not $\z^n$ itself or the structure of the function $c$. Directly learning the reward function requires a more complex class of predictors to capture this relationship compared to an end-to-end method. Simplifying the predictor class $r$ is crucial as the complexity of $r$ directly impacts the difficulty of solving $\max_{\v\in \mathcal{V}} r(\v,\x)$ in large/continuous action space. Our approach allows for simpler model classes while still capturing the same complexity in modeling $\mathbb{E}_{\z \sim Z|\v,\x}[c(\v,\z)]$. We further see this explicitly in the numerical computations, see section \ref{sec:experiments}.

\newpara{Distinction from exogenous case}
Finally, we describe and compare against the setting under exogenous random variables. We will also use this methodology in conjunction with ours in the two stage information-gathering setting which we will introduce in section \ref{sec:info}. The first stage is endogenous, while the second is exogenous.  For ease of comparison, we denote all parameters in the exogenous case with a bar. Here one observes features ${\x}$ and corresponding realizations of uncertainty $\bar{\z}$ coming from a distribution $\bar{\Z}|\x$ that depends only on ${\x}$. On the other hand, in the endogenous case, $\z$ depends on both $\x$ and $\v$. The objective in the exogenous case is 
\begin{equation}
\label{eq:exo-obj}
    \max_{\v \in \mathcal{P}} \mathbb{E}_{\bar{\z} \sim \bar{\Z}|\x}[c(\v, \bar{\z})].
\end{equation}
Computing this expectation is difficult. So, one goal of an end-to-end approach is to replace this with a point forecast which produces the same decision and objective value. The objective is to learn a point forecast $\bar{f}(\x)$ to replace the distribution $\bar{Z}(\x)$. Replacing $\bar{Z}(x)$ with the deterministic $ \bar{f}(\x) $ in problem \eqref{eq:exo-obj} give us $\max_{\v \in \mathcal{V}} c(\v, \bar{f}(\x))$ which is computationally much simpler to solve. 
Given a point forecast $\z = \bar{f}(\x)$ denote the optimal corresponding decision by $v^*(\z)$:
\begin{equation}
    v^*(\bar{\z}) = \arg\max_{\v\in\mathcal{V}} c(\v,\bar{\z}).
\end{equation}
We would like to learn an $\bar{f}$ for which 
\begin{equation}
\label{eq:exo}
   c(v^*(\bar{f}(\x)), \bar{\z}) \approx \max_{\v \in \mathcal{V}} \mathbb{E}_{\bar{\z} \sim \bar{\Z}|\x}[c(\v, \bar{\z})]
\end{equation}
which allows us to replace $\bar{\Z}|\x$ with $\bar{f}(\x)$ in \eqref{eq:exo-obj}. 
This is similar to our proposed objective \eqref{eq:expectation}.  
Under exogenous variables, the common data-driven objective is to learn a model $\bar{f}$ which maximizes the reward/objective of the corresponding decisions $v^*(\bar{f}(\x))$ that it takes.  See for example \cite{elmachtoub2022smart}. 
Given data $({\x}^n, \bar{\z}_n)_{n=1}^N$, we wish to solve
\begin{equation}
\label{eq:exo-train}
    \hat{f}_{\text{exo}} = \arg \max_{f} \sum_{n=1}^N c(v^*(\bar{f}({\x}_n)), \bar{\z}_n).
\end{equation}
After learning some $\hat{f}_{\text{exo}}$ (for the exogenous case) one then takes decisions $w^*(\hat{f}_{\text{exo}}(\x))$.  In the endogenous setting, we predict a point statistic of the uncertainty so that the predicted reward for any action  taken (including historical ones) are close to their realization. In contrast, here in the exogenous case, we are predicting a statistic of the uncertainty so that the value of the optimal decision given a point forecast is close to the optimal expected reward. The methodology used for the exogenous case cannot be applied to the endogenous case because it would require one to have access to counterfactual information. The objective value of a decision $w^*(\hat{f}(\x))$ 
cannot be evaluated because it was not taken historically, and we hence we do not have any information on the random variable $\z$ that would depend on the decision. In the exogenous case, $\z$ does not depend on the decision, so we can indeed evaluate the decision cost by using historically observed $\z$.

\subsection{Algorithms} 
\label{sec:algorithms1}

We turn to an important discussion on the challenges and the potential methods for solving the learning problem presented in \eqref{eq:end-to-end}. The main difficulty in solving \eqref{eq:end-to-end} is its non-convexity. Note that each term term $(c(\v_n, \z) - c(\v_n, \z_n))^2$ is a non-convex function of $\z$ whenever $c(\v,\z)$ is non-linear in $\z$. We first present an exact integer optimization-based formulation for solving the non-convex problem, then a more efficient sampling-based approximation. In the former exact case, we will assume that $c$ is piece-wise linear and convex, while in the latter we make no such assumption. 

\newpara{Exact reformulation:} We first formulate \eqref{eq:end-to-end} as a mixed-integer quadratic optimization problem. Let $c$ be the maximum of $K$ linear functions $c^1, \dots, c^K$ so that 
 $   c(\v, \z) = \max_{k=1,\dots,K} c^k(\v,\z) $. The examples in section 2 have this structure for instance. The formulation is
\begin{align}
    &\ \min_{f,v,y} \sum_{n=1}^N s_n \quad \text{subject to} \label{eq:solve-exact} \\
    &\ s_n \geq (c^k(\v_n, f(\x_n, \v_n)) - r_n)^2 - M(1 - y_{n,k}) \nonumber\\
    &\ c^k(\v_n, f(\x_n, \v_n)) \geq c^j(\v_n, f(\x_n, \v_n)) - M(1-y_{n,k}) \nonumber \\ 
    &\ \sum_{k=1}^K y_{n,k} = 1, \forall n, \quad \text{and} \quad y_{n,k} \in \{ 0,1 \} \nonumber
\end{align}
where for ease of notation $r_n$ is a constant equal to the $n^{th}$ datapoint's objective value: $c(\v_n, \z_n)$. The binary variable $y_{n,k}$ is forced to equal $1$ for any $f$ such that $c(\cdot,\cdot)$ is equal to $c^k(\cdot,\cdot)$. We have $c^k(\v_n,f(\x_n,v_n)) \geq c^j(\v_n,f(\x_n,\v_n))$ for all $j = 1, \dots, K$ for exactly one index $k$, hence we can set $y_{n,k} = 1$ and the constraints hold. For every other $k$, the constraints do not hold. However, since $y_{n,k} = 0$ the constraint $ c^k(w_n, f(\x_n, \v_n)) \geq c^j(\v_n, f(\x_n, \v_n)) - M $ do hold for large enough $M$. Finally, we force exactly one $y_{n,k}$ to equal 1 by $\sum_{k} y_{n,k} = 1$. 

Finally, $s_n$ is simply equal to $(c^k(\v_n, f(\x_n,\v_n)) - r_n)^2$ for the appropriate $k$ where $g^k=g$. Indeed, for $y_{n,k} = 1$, the first constraint becomes equivalent to $s_n \geq (c(\v_n, f(\x_n,\v_n))-r_n)^2$. Since the objective function is to minimize $\sum_{n} s_n$, it follows that $s_n$ will take the smallest possible value which will be equal to the maximum of all $c((\v_n, f(\x_n,\v_n))-r_n)^2 - M(1-y_{n,k})$. Whenever $y_{n,k} = 0$, the constraints can essentially be ignored since they impose a smaller lower bound. So the maximum is achieved at $k$ for which $y_{n,k}=1$, making $s_n =  (c(\v_n, f(\x_n,\v_n))-r_n)^2$. 

If $f(\x, \v)$ were a linear function, then the above formulation is a mixed integer quadratic-convex 
optimization problem and can be solved by off-the-shelf solvers. Of course, one can use augmented features and kernel functions to increase the expressivity of the prediction model while remaining linear.

While exact, the formulation presented in the previous subsection is intractable as the amount of data increases. Here we provide two sampling-based approaches that are computationally more efficient, albeit do not guarantee optimality.

\newpara{Perturbing the mean-squared predictor:}
(1) First compute the two-stage approximator (by solving $\min_{\theta} \sum (f_\theta(w_n,x_n) - z_n)^2$)
    and let $\hat{\theta}$ be the weights found.
    (2) For each sample $s = 1, \dots, S$, perturb the weights $\hat{\theta}$ by some random gaussian noise $\delta^s$ to produce a sample $\theta^s = \hat{\theta} + \delta^s$. 
    (3) Perform gradient descent using each sample $\theta^s$ as an initialization point. Choose the model with the best in-sample \emph{task-based} loss. 

\newpara{Iterative learning:} (1) First, create new smaller datasets, with the $k^{th}$ one containing the first $k \cdot D$ datapoints $(x_1, w_1, z_1), \dots, (\x_{k \cdot D}, \v_{kD}, \z_{kD})$ where $k=1,\dots,K=N/D$.  
    (2) For $k=1$, apply the previous method by sampling from perturbing the mean-squared predictor. Let $\theta^1$ be the final model.
    (3) For $k > 1$, use $\theta^{k-1}$ as the initial model, generate $S$ new samples by perturbing $\theta^{k-1}$, then apply gradient descent to minimize task-based loss. Choose $\theta^k$ with the best loss from these. Finally,
    (4) return the model with $\theta^D$.

See Algorithm \ref{e2e-endo-alg} for a summary. The iterative learning method is essentially a super-set of the first method. We observe it provides generally better results as well in the numerical experiments. Intuitively, this makes sense: as we add more data, we fine-tune the previous model learned. Moreover, this is also useful when data arrives online. One observes data up to a time point, then make a new decision, then observe the outcome. This new observation becomes a new data point that can be used for training. 

\begin{algorithm}[t]
\caption{Endogenous end-to-end learning and decision making pipeline}\label{e2e-endo-alg}
 \hspace*{\algorithmicindent} \textbf{Input:} Training data $\{(\x^n , \z^n, \v^n) \}_{n=1}^N$, out-of-sample $\x$ \\
 \hspace*{\algorithmicindent} \textbf{Output:} Decision $\v$
\begin{algorithmic}
\STATE Learn point forecast $f(\x, \v)$ by solving \eqref{eq:end-to-end}.
\STATE $ \quad $ If objective $c$ piece-wise linear, solve by exact method (see \eqref{eq:solve-exact}).
\STATE $ \quad $ Else, solve by sampling method.
\STATE For out-of-sample $\x$, take decisions by solving \eqref{offline-dec}.
\STATE $ \quad $ If $\mathcal{V}$, small, solve by enumerating all $\v \in \mathcal{V}$.
\STATE $ \quad $ Otherwise solve by gradient descent, or traditional optimization methods. 
\end{algorithmic}
\end{algorithm}

\subsection{Generalization}
\label{sec:theory}

Here we analyze two theoretical properties of our proposed method. First we show that the idealized goal in \eqref{eq:expectation} is achievable. We show that such a function $f^*$ exists so that $c(\v, f^*(\v,\x)) = \mathbb{E}_{\z \sim Z|\v,\x} [c(\v,\z)]$ given exact knowledge of the distribution $Z|\v,\x$. We answer this under some mild assumptions in Proposition \ref{prop:exist}. Next, we prove generalization bounds in Theorem \ref{thm:generalization} to show how much data is needed to learn $\hat{f}_{\text{end-to-end}}$.
\begin{proposition}
\label{prop:exist}
    For continuous objective functions $c(\v, \z)$ with respect to $\z$, there exists $\hat{\z}$ in the convex hull of the support of $Z|\v, \x$ so that 
    \begin{equation}
        c(\v, \hat{\z}) = \mathbb{E}_{\z \in Z|\v, \x} [c(\v, \z)].
    \end{equation}
\end{proposition}
If the class of functions $f$ from is expressive enough, it is possible to achieve \eqref{eq:expectation}. Next we focus on generalization bounds. To do so, we use a common technique using Rademacher complexity which we will define formally in Definition \ref{def:rademacher}. Intuitively,  as the complexity, as measured by Rademacher complexity, of the hypothesis class of functions to choose $f$ from increases, often the more data it takes to generalize out of sample. 

\begin{theorem}
\label{thm:generalization}
     For any function $f \in \mathcal{F}$, we define out-of-sample error/loss $l$ and the empirical loss $\hat{l}$ over a random sample of $N$ datapoints $(\v_n,\x_n,\z_n), n=1,\dots,N$  
     \begin{align}
      l(f) = \mathbb{E}_{\v,\z}[\paren{c(\v, f(\x, \v)) - c(\v, \z)}^2], \quad
      \hat{l}(f) = \sum_{n=1}^N (c(\v_n, f(\x_n, \v_n)) - c(\v_n, \z_n)))^2. \nonumber
     \end{align}
     For any $L$-Lipschitz function $c$ (with respect to $\z$), $c(\v, \x) \in [0,1] \forall \v \in \mathcal{V}$ and all $\z$ in the support of $Z|\v,\x$, then we have with probability $1 - \delta$,
     \begin{equation}
         l(f) \leq \hat{l}(f) + 2L\sqrt{2} \mathscr{R}_N(\mathcal{F}) + \paren{\frac{8\log 2/\delta}{N}}^2.
    \end{equation}
\end{theorem}
For many hypothesis classes $\mathcal{F}$, we can bound $\mathscr{R}_N(\mathcal{F})$ by a term that converges to $0$ as $N\to\infty$ and at a rate $O(1/\sqrt{N})$ for common function classes like linear functions. See for example \cite{bartlett2002rademacher}. So, the overall generalization gap decreases at a $O(1/\sqrt{N})$ rate. 

\begin{definition}[Multidimensional Rademacher Complexity]
\label{def:rademacher}
 The empirical Rademacher complexity of the hypothesis class of functions $\mathcal{F} : (\x, \v) \to \R^d$  is given by
$    \mathscr{R}_N(\mathcal{F}) = \mathbb{E}_{\{(\v^n, \x^n)\}_{n=1}^N} \mathbb{E}_{\sigma}\left[ \sup_{f \in \mathcal{F}} \frac{1}{N} \sum_{n=1}^N \sum_{k=1}^{d} \sigma_{nk} f_k(\x_n, \v_n) \right]$
where $\sigma_{nk}$ are i.i.d. variables uniformly sampled from $\{-1,1\}$ (also known as Rademacher variables). 
\end{definition}

\section{Endogenous robust end-to-end learning}
\label{section:robust}
Since we have limited data, there will always be error in any function we learn to approximate $f^*$ as in \eqref{eq:expectation}. Indeed, there are many models that we can learn that approximately minimize the task-loss from Definition \ref{def:task-loss}. Each such model can lead to very different decisions. We propose a robust optimization and learning method that considers all such possible models. 

Suppose that we know $f^*$ belongs to some uncertainty set of functions $\mathcal{U}_{\epsilon}$, where ${\epsilon}$ represents the ``size'' of the set, and intuitively how robust we wish to be. 
Then we wish to make a decision which maximizes its objective with respect to all of the functions in $\mathcal{U}_{\epsilon}$. This ensures at least a lower bound on the expected objective of our final decision. Hence, we take decision
\begin{equation}
\label{eq:offline}
     v^*_{\text{robust}}(f; \x) = \arg\min_{\v \in \mathcal{V}} \max_{f \in \mathcal{U}_{\epsilon}} c(\v, f(\x, \v)).
\end{equation}
Next, we will propose a construction for the uncertainty set $\mathcal{U}_{\epsilon}$. We select a set that comprises the set of models whose task-based loss on the data is bounded by some user-decided threshold. 
We propose setting $\mathcal{U}_{\epsilon}$ to be the set of functions in some function class $\mathcal{F}$ (linear for instance) which are consistent with the data up to some error $\epsilon$:
\begin{equation}
    \mathcal{U}_{\epsilon} = \left\{ f \in \mathcal{F} : \mathcal{E}(f) \leq \beta + \epsilon \right\}
\end{equation}
where $\beta$ is the error of the optimal approximator given only data: $\beta = \min_{f \in \mathcal{F}} \mathcal{E}(f)$. Again, $\mathcal{E}(f)$ denotes the task-based loss of $f$ as defined in \eqref{def:task-loss}.


\begin{remark}
Note that it is crucial that the uncertainty set is defined in terms of the error in the cost of $c$ itself. If instead one focuses only on error in prediction $\sum (f(\x_n, \v_n) - \z_n)^2$ then $f$ would target predicting $\mathbb{E}[\z]$. However, in general, predicting the mean is incorrect as $\mathbb{E}[c(\v,\z)] \not= c(\v,\mathbb{E}[\z])$ when $c$ is non-linear in $\z$.    
\end{remark}

Before we state the theorem, we discuss the impact of the class of functions $\mathcal{F}$. If for example $\mathcal{F}$ is parameterized by a larger number of variables, then the size of the uncertainty set should intuitively increase as well. We make this formal by introducing the common notion of \emph{coverings}: 
\begin{definition}
    Given a set $\mathcal{F}$ with a metric $d$, a $\gamma$-covering is any set $F$ for which any element of $\mathcal{F}$ is within a distance of $\gamma$ from some element of $F$. That is, $F$ is a $\gamma$-covering if 
\begin{equation}
     \forall f \in\mathcal{F}, \exists \tilde{f} \in F \ \text{s.t. } d(f,\tilde{f}) \leq \gamma 
\end{equation}
A $\gamma$-\emph{covering} $\mathcal{N}_\gamma$ is the smallest $\gamma$-covering:
    \begin{equation}
        \mathcal{N}(\mathcal{F}, d, \gamma) = \inf \{ |F| : F \text{ a $\gamma$-cover of } \mathcal{F} \}
    \end{equation}
\end{definition}
In general, we will consider metrics over parameterized functions $f_\theta$ and use $l_2$ norm: 
\begin{equation}
    d(f_\theta, f_{\tilde{\theta}}) = \norm{\theta - \tilde{\theta}}.
\end{equation}
We will consider function classes $\mathcal{F}$ that correspond to linear functions or more general neural networks. So, we can indeed parameterize them as $f_\theta$ where $\theta$ then correspond to the weights. We make the final assumption on the smoothness of the hypothesis class.
\begin{assumption}
\label{assumption2}
    The class of predictive functions $f_\theta(\x,\v)$ for any given $w$ and $x$ is $M$-Lipschitz with respect to $\theta$. That is, for any $\theta, \theta'$
    \begin{equation}
        |f_{\theta}(\x, \v) - f_{\theta'}(\x, \v)| \leq M \norm{\theta - \theta'}_2.
    \end{equation}
\end{assumption}
\begin{theorem}
\label{theorem:uncertainty}
Suppose data $(x_n, w_n, z_n)_{n=1}^N$ is generated as in assumption \ref{ass:data}. Choose $f^* \in \mathcal{F}$ to be the ideal function as in \ref{eq:expectation}. However, in general the hypothesis class $\mathcal{F}$ is not expressive enough to achieve this exactly. So, let 
\begin{equation}
    f^* = \arg\min_{f\in\mathcal{F}} \mathbb{E}_{\x}\mathbb{E}_{\z\sim \Z|\x, \v} \left[ (c(\v, f(\x, \v)) - c(\v, \z))^2 \right]
\end{equation}
and let $\Delta$ be the error of $f^*$ as defined above. Then, choose
\begin{equation}
    \epsilon_N = \frac{8}{N} \log \paren{\frac{|\mathcal{N}_{\gamma}|}{\delta}} + 3M\gamma + \Delta
\end{equation}
where $\mathcal{N}_\gamma$ is the smallest $\gamma$-covering of $\mathcal{F}$ and $M$ is the Lipschitz constant of assumption \ref{assumption2}. Then with probability $1 - \delta$ we have that $f^* \in \mathcal{U}_{\epsilon_{N}}$.
\end{theorem} 
For linear hypothesis classes, we can analytically compute the covering number and optimize over $\gamma$. We leverage this in the proof of above theorem in appendix \ref{sec:proofs} to obtain the following corollary. 
\begin{corollary}
To optimize over the space $\mathcal{F}$ of $d$-dimensional linear functions with norm bounded by $1$, one can set $\epsilon = O(\frac{1}{N}(d \log N + \log 1/\delta + M))$ in Theorem \eqref{theorem:uncertainty}. 
\end{corollary}

\subsection{Algorithmic Methods}
\label{sec:alg}
We now focus on solving the problems \eqref{eq:offline}. We can split the problem in two: the inner problem
\begin{align}
\label{eq:inner}
G_{\text{inner}}(\x,\v) =&\ \max_{f \in\mathcal{U}}  c(\v, f(\x, \v))
\end{align}
and the outer problem $\min_{\v \in \mathcal{V}} G_{\text{inner}}(\x, \v)$ to finally make decisions.

One approach to solving the above problem is by removing the constraint $f \in \mathcal{U}_{\epsilon}$ and adding a regularization term in the objective to penalize violating it. This is similar to the approach in \cite{xie2021bellman} introduced for robust reinforcement learning. Converting this method into an end-to-end approach would give
\begin{equation}
\label{eq:reg}
\max_{f} c(\v, f(\x, \v)) - \lambda \cdot \mathcal{E}(f).     
\end{equation}

However, this approach would require tuning $\lambda$ for every value of $\v$. This becomes computationally expensive as we would need to do this for $\v$ as we solve the outer problem. Alternatively, we can remain with a single $\lambda$ that may be suboptimal overall.
We instead propose a method that requires no such tuning other than choosing $\epsilon$, the size of the uncertainty set. Our proposed algorithm consists of two alternating steps. As long as the prediction error on the data, namely $\mathcal{E}(f)$, is below the threshold $\epsilon$, we only focus on maximizing $\max_{f} c(\v, f(\x, \v))$ (with no constraints on $f$). Once the constraint is violated, that is the error on the data is $\mathcal{E}(f) > \epsilon$, we only focus on minimizing $\min_{f} \mathcal{E}(f)$ until the error threshold is met and $\mathcal{E}(f) \leq \epsilon$ again. We then repeat the process. This is outlined again in Algorithm 
\ref{alg:inner}. Intuitively, this would be more stable than the regularization approach since it always guarantees the constraint $f \in \mathcal{U}_\epsilon$ is enforced and there is no additional hyperparameter to tune. We compare the two methods computationally on an assortment optimization problem in section \ref{sec:ass} and explicitly see the advantage in performance of our approach.

\begin{algorithm}[t]
\caption{Solving the inner problem \eqref{eq:inner}.}
\label{alg:inner}
 \hspace*{\algorithmicindent} \textbf{Input:} $\v, \x, \epsilon, T$ \\
 \hspace*{\algorithmicindent} \textbf{Output:} $f$
\begin{algorithmic}
\FOR{T epochs}
\STATE IF {$\mathcal{E}(f) > \epsilon$}: Perform gradient descent on $\min_{f} \mathcal{E}(f)$
\STATE ELSE: Perform gradient descent on $\max_{f} c(\v, f(\x, \v))$
\ENDFOR
\end{algorithmic}
\end{algorithm}

\newpara{Solving the outer problem.} We now consider solving the outer problem $\max_{\v\in\mathcal{V}} G_{\text{inner}}(\x,\v)$. If $\mathcal{V}$ is discrete and small enough, one can simply compute $G_{\text{inner}}(\x,\v)$ directly for each decision. In this section we consider the case that $\mathcal{V}$ is too large to enumerate (and is either a discrete, or continuous space). We propose to solve this problem by a cutting plane approach. We construct a sequence of sets $\mathcal{U}^{(0)}, \mathcal{U}^{(1)}, \dots$ as follows. Initially, $\mathcal{U}^{(0)}$ is empty. We successively add new predictors $\tilde{f} \in \mathcal{U}$ to each set $\mathcal{U}^{(k)}$ as follows. Given that $\mathcal{U}^{(k)}$ contains elements $f^{(1)}, \dots, f^{(k)} \in \mathcal{U}$, then we find the optimal solution $\v^{(k)}$ using this uncertainty set:
\begin{equation}
\label{eq:cutting-plane}
    \v^{(k)} = \arg \min_{\v \in \mathcal{V}} \max_{j=1,\dots,k} c\paren{\v, f^{(j)}(\v, \x)}.
\end{equation}
In order to find a \emph{cut}, it suffices to find a function $\tilde{f} \in \mathcal{U}$ which produces an upper predicted objective for $\v^{(k)}$ than $f^{(1)}, \dots, f^{(k)}$. That is, $\tilde{f}$ such that
\begin{equation*}
    c\paren{\v^{(k)}, \tilde{f}(\x, \v^{(k)})} \geq \max_{j=1,\dots,k} c\paren{\v^{(k)}, {f^{(j)}}(\x, \v^{(k)})}.
\end{equation*}
This can be done by simply solving $G_{\text{inner}}(\x, \v^{(k)})$. The algorithm as a whole is then given in Algorithm \ref{cutting-algorithm}. Note that $\max_{j=1,\dots,T} c\paren{\v^{(T)}, {f^{(j)}}(\x, \v^{(T)})}$ is an upper bound for the objective $\min_{\v \in \mathcal{V}} \max_{f \in \mathcal{U}} c(\v, f(\x,\v))$ whereas $G_{\text{inner}}(\x, \v^{(T)})$ is a lower bound. 

\begin{algorithm}
\caption{Cutting plane for outer problem \eqref{eq:offline}.}\label{cutting-algorithm}
 \hspace*{\algorithmicindent} \textbf{Input:} Training data $\{(\x^n , \z^n, \v^n) \}_{n=1}^N$, out-of-sample $\x$, uncertainty set size $\epsilon$ \\
 \hspace*{\algorithmicindent} \textbf{Output:} Decision $\v$
\begin{algorithmic}
\STATE Set $\mathcal{U}^{(0)} = \emptyset$ and set $\v^{(0)}$ to be any element of $\mathcal{V}$.
\FOR{$k=0,\dots$}
    \STATE Compute $G_{\text{inner}}(\x, \v^{(k)})$, let $\tilde{f}$ be the minimizer found. 
    \IF{$G_{\text{inner}}(\x, \v^{(k)}) > c\paren{\v^{(k)}, {f^{(j)}}(\x, \v^{(k)})}, \forall j \leq k$}
    \STATE Set $\mathcal{U}^{k+1} = \mathcal{U}^{(k)} \cup \{ \tilde{f} \}$
    \ELSE 
    \STATE Finish and return optimal $\v^{(k)}$
    \ENDIF
    \STATE Compute $\v^{(k+1)}$ by solving \eqref{eq:cutting-plane}.
\ENDFOR
\end{algorithmic}
\end{algorithm}

Therefore, at each iteration of the cutting-plane algorithm we can calculate an optimality gap, the ratio between this upper and lower bound. One can then terminate the cutting plane algorithm once this gap reaches some pre-specified threshold. Whenever $G_{\text{inner}}(\x,\v)$ is convex in $\v$,  this algorithm will converge to a global optimum \cite{kelley1960cutting}. Otherwise, it would converge to a local minima. In this case, one can sample different initial decisions $\v^{(0)}$ and run the cutting plane algorithm above for each, and finally choose the best decision found. 

In summary, we have presented a tractable robust learning variant of the original end-to-end approach from section \ref{sec:framework}. We have constructed an uncertainty set over possible predictive models, proved this set contains the true $f^*$ with high likelihood, and presented a cutting plane algorithm to solve the min-max optimization problem. 


\section{Application to information gathering usecase: combining exogenous and endogeneous end-to-end approaches}
\label{sec:info}

Here we consider a novel application of the end-to-end method to a class of 2-stage optimization problems with information-gathering.  As an example, consider a multi-warehouse inventory allocation problem. In the first stage, we can choose a single location to poll to learn the demand for the next time period. In the second stage, we must (1) predict the demand at all other locations, conditioned on our previous observation from the poll then (2) decide how much inventory to allocate at all warehouses.  We note that this is different from traditional 2- or multi-stage stochastic optimization problems. There, the first stage is some operational decision (for e.g. inventory allocation in a warehouse), and then in the second stage some additional information is revealed (such as demand). In our setting, there is a deliberate initial action to decide which additional information to reveal ahead of time (for e.g. before allocating inventory, we can poll one location to know exact demand). 

One approach would be to learn a model that predicts, for every location, the expected cost that results from polling it. This simplifies the problem but hides the structure behind it and makes the learning problem more complex, requiring a richer class of functions to approximate it. We explicitly observe the advantage of using the problem structure while learning an end-to-end model in the experiments. 

Formally, we are given exogenous random variables $\z = (z_1, \dots, z_d)$, independent of the decisions that we take. In the first stage, the decision space will consist of choosing some index $w \in \mathcal{W} = \{1, \dots, d\}$ to survey, or gain more information about, the $w^{th}$ entry of $\z$, namely $z_w$. This could be more general beyond observing a single value, for example observing multiple values. We will see this in an experiment in section \ref{sec:electricity}. But we will keep this simple here for the sake of notation. 
As an example, this could correspond to setting up a survey to learn more about the demand of the $w^{th}$ 
product. In the second stage, we make a prediction for the remaining random variables, conditioned on the observation of $z_w$. Note this prediction does depend on the decision we initially took to survey the $w^{th}$ random variable. In the second stage, we are given some auxiliary decision variables $\mathbf{v} \in \mathcal{V}$ with objective function $c(\mathbf{v}, \z)$. In short, the entire process is as follows:
\begin{enumerate}
    \item For an out-of-sample point $\x$, we make a decision $w$ to observe $z_w \sim Z_w|\x $. 
    \item Given the observation $z_w$, the full vector $\z$ is distributed according to $Z|\x, Z_w=z_w$.
    \item We are now given some second-stage decision-making problem with variables $\mathbf{v} \in \mathcal{V}$ with objective $c(\mathbf{v},\z)$ and we wish to make decision $\mathbf{v}$ minimizing expected cost: 
    \begin{equation}
        \label{eq:second-stage}
        \min_{\mathbf{v} \in \mathcal{V}} \mathbb{E}_{\z \sim Z|\x, {Z_w = z_w}} [c(\mathbf{v},\z) ].
    \end{equation}
    \item Ultimately, we wish to know which observational decision $w$ will minimize  overall loss: 
    \begin{equation}
    \label{eq:info}
        \min_{w \in \mathcal{W}} \mathbb{E}_{z_w \sim Z_{w}(\x)} \left[ \min_{\mathbf{v} \in \mathcal{V}} \mathbb{E}_{\z \sim Z|\x, {Z_w = z_w}} [c(\mathbf{v}, \z) ] \right].
    \end{equation}
\end{enumerate}
In terms of data, we observe $N$ points $(\x^n, w^n, \z^n), n=1\dots,N$ where $\z^n$ is distributed according to an (unknown) distribution $Z|\x^n$. Given decision $w^n$, we observe the realization of $\z^n_{w^n}$ before making the second-stage decision.  To train we proceed as follows. We first begin by simplifying the inner expectation in \eqref{eq:second-stage}. After making a decision $w$ of observing $z_w$, we can make a forecast for some statistic of the distribution $Z|\x, {Z_w = z_w}$. Let $p(\x, z_w)$ denote this prediction for all of $\z$ conditioned on observing $z_w$ as well as features $\x$. In particular, we are now in a similar setting as the traditional end-to-end problem. For example, given forecast $p(\x, z_w)$ for product demand, we then need to solve an optimization problem to optimize the inventory allocation. That is, we take decision
\begin{equation}
    v^*(p(\x, z_w)) = \arg\min_{\mathbf{v} \in \mathcal{V}} c(\mathbf{v}, p(\x, z_w)).
\end{equation}
Essentially, we will learn these point forecasts $p(\x, z_w)$ in order to remove the expectations from problem \eqref{eq:info}.
This is similar to the traditional end-to-end framework in \eqref{eq:exo}. 
\begin{enumerate}
    \item We first learn $p(\x, z_w)$ to predict $\z$ conditioned on observing $z_w$ for the $w^{th}$ random variable. We want such a $p$ to approximate \eqref{eq:second-stage}. That is, we need 
    \begin{equation}
    \label{eq1}
    c(v^*(p(\x, z_w)), \z) \approx \min_{\mathbf{v} \in \mathcal{V}} \mathbb{E}_{\z \sim Z(\x)|_{Z|\x, {Z_w = z_w}}} [c(\mathbf{v},\z) ].
    \end{equation}
We learn such a $p$ by solving the following empirical risk minimization problem, similar to \eqref{eq:exo-train}:
\begin{equation}
    \label{eq:exo-obj1}
    \min_{p} \sum_{n=1}^N c(v^*(p(\x^, z^n_{w^n})), \z^n).
 \end{equation}
\item Now, substituting \eqref{eq1} into \eqref{eq:info} our final problem simplifies to
\begin{equation}    \min_{w} \mathbb{E}_{z_w \sim Z_w(
\x)} [ c(v^*(p(\x, z_w)), \z) ].
\end{equation}
This problem now falls under our end-to-end framework with endogenous random variables because the objective function depends on $p(\x, z_w)$ which in turn depends on the first-stage decision $w$. So, similar to \eqref{eq:expectation}, we wish to learn a single point forecast $f(w, \x)$ to replace the expectation over $\z$. That is, our goal is to learn a function $f$ so that 
\begin{equation}
    c\Big(v^*\big(p(\x, f_w(w, \x))\big), f(w, \x)\Big) \approx \mathbb{E}_{z_w \sim Z_w|\x} [ c(v^*(p(\x, z_w)), \z) ].
\end{equation}
We replace $\z$ with a point forecast $f(w, \x)$. To learn $f$ we use a version of our  method in \eqref{eq:end-to-end}:
\begin{equation}
\label{opt:endo1}
    \min_{f} \sum_{i=1}^n \Big( c\Big(v^*\big(p(\x^i, f_{w^i}(w^i, \x^i))\big), f(w^i, \x^i)\Big) - c\Big(v^*\big(p(\x^i, z_{w^i}^i))\big), \z^i\Big) \Big)^2.
\end{equation}
\item Finally, for an out-of-sample $\x$, we make decisions by solving 
\begin{equation}
\label{eq:opt-info}
    \arg\min_{w} c\Big(v^*\big(p(\x, f_w(w, \x))\big), f(w, \x)\Big).
\end{equation}
\end{enumerate}
In practice, we cannot observe $z_w$ before making decision $w$, so in \eqref{opt:endo1} and \eqref{eq:opt-info} we ``observe'' the $w^{th}$ entry of the predicted $f(w, \x)$ instead.
Algorithm \ref{e2e-alg-info} provides a concise description. We use the first end-to-end approach from section \ref{sec:framework}, although the robust variant can also be used to solve the second stage.

\begin{algorithm}[t]
\caption{End-to-end information-gathering}\label{e2e-alg-info}
 \hspace*{\algorithmicindent} \textbf{Input:} Training data $\{(\x^n , \z^n, \v^n) \}_{n=1}^N$, out-of-sample $\x$ \\
 \hspace*{\algorithmicindent} \textbf{Output:} First-stage decision $w$ and second-stage decision $\v$.
\begin{algorithmic}
\STATE Learn model $p(\x, z_w)$ to predict $\z$ conditioned on observing $z_w$. Learn $p$ by solving \eqref{eq:exo-obj1} with gradient descent. 
\STATE $ \quad $ Compute gradients $\partial v^*(\z)/\partial\z$ using any exogenous end-to-end approaches from prior work such as \citep{donti2017task, cristian2023end,amos2017optnet}.
\STATE Learn point forecast $f(w, \x)$ by solving \eqref{opt:endo1} by gradient descent.
\STATE For out-of-sample $\x$, choose decision $w$ by solving \eqref{eq:opt-info}.
\STATE For out-of-sample $\x$ and decision $w$, observe $z_w$. And take second-stage decision $v^*(p(\x, z_w))$.
\end{algorithmic}
\end{algorithm}

\section{Computational experiments}
\label{sec:experiments}

\subsection{Pricing}
 
We first consider the case of pricing a single item, but the demand is also a function of additional feature information $\x$. Given a price vector $\v$ for $K$ items and a corresponding demand $\z$, the objective is $c(\v,\z) = \v^T\z$. 

s\newpara{Experiment setup:} The feature data $\x$ is generated at random from a $d=25$-dimensional normal distribution. Then, the demand is generated by two components: first a baseline dependent only on a quadratic function of the features plus a second component dependent on a polynomial function of the price. Demand is distributed normally, with variance 1. The mean demand of each item $k$ is denoted by $\bar{\z}_k$ which is determined by $\bar{z}_k = \lambda \cdot (B^T\x)^2_k + (1 - v_k)^p$ where $\lambda$ is a constant to determine the relative weight between the effect of price and the other features. These components are meant to create realistic price-demand interactions. For example, as price increases, demand is expected to decrease, with $p$ describing how strong this effect is. Moreover, note this only describes the mean demand. There is inherent noise which we will model as Gaussian.
For the main experiments, we use $n=2000$ training data points. We construct such a dataset $(\x_n, \v_n, \z_n), n=1,\dots,N$.

\newpara{Benchmarks}
We compare against a two-stage approach (which we will refer to as a baseline). This predicts the demand as a function of price and features.  That is, it solves $\hat{f} = \min_{f \in \mathcal{F}} \sum_{n=1}^N (f(\x_n, \v_n) -  \z_i)^2.$ Then for an out-of-sample $\x$, one chooses price $\v$ with $\arg\min_{\v} c(\v, \hat{f}(\x, \v))$. We use a linear model class for $\mathcal{F}$. We also compare against a purely reward-learning approach, which instead trains a model $h$ to predict revenue directly $h(\x, \v)$: $\min_{h} \sum_{n=1}^N (h(\x_n, \v_n) - c(\v_n, \z_n))^2$ and takes actions $\arg\min_{\v\in\mathcal{V}} h(\x, \v)$. For the reward-learning method, we use a neural network with a hidden layer of width 100. We also consider a Gaussian Process regression method with RBF kernel and a kernel (WhiteKernel in the python scikit-learn package) to explain and capture the noise $(\alpha)$ of the signal as independently and identically normally-distributed. It can be interpreted as the variance of additional Gaussian measurement noise on the training observations.

\begin{figure*}[t]
    \centering
\includegraphics[width=1\linewidth]{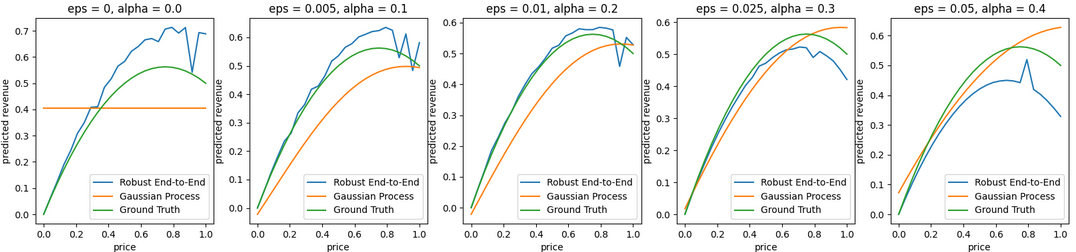}
    \caption{Evaluation of inner lower bound for each price at various choices of robustness parameters $\epsilon$ and noise parameter $\alpha$ for the Gaussian process method.
    }
    \label{fig:eps-effect}
\end{figure*}

\newpara{Effect of robustness parameter $\epsilon$.} First, to gain some intuition, we illustrate the effect of $\epsilon$. For each price decision $\v$, we evaluate the robust lower bound $\min_{f \in \mathcal{U}_\epsilon} c(\v, f(\x,\v)) $. We plot this prediction for all prices $\v$ for several values of $\epsilon$ in Figure \ref{fig:eps-effect} (here we fix a single $\x$ for illustration). We do the same for the Gaussian Process, and compare against various levels of noise parameters for it. For our robust end-to-end approach, since the data is noisy, we learn an imperfect model of $f$ at $\epsilon = 0$. As we increase robustness ($\epsilon$), we solve the problem $\min_{f \in \mathcal{U}_\epsilon} c(\v, f(\x, \v))$. So, the prediction for revenue at each price $\v$ slowly becomes more robust to noise in the data and eventually becomes a true lower bound to the true revenue. See the blue curve in Figure \ref{fig:eps-effect} illustrating this as $\epsilon$ is increases from the left to right plot. On the other hand, the noise parameter for the Gaussian process method does not equate to robustness, varying the $\alpha$ parameter can only improve the method's fit. 

\begin{figure}[t]
\centering    \includegraphics[width=0.8\linewidth]{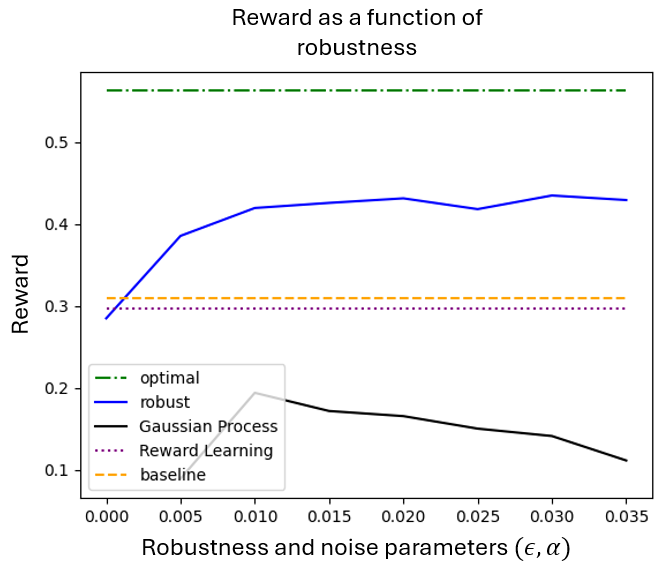}
    \caption{Decision-revenue across different robustness/noise parameters for pricing problem.}
    \label{fig:revenue}
\end{figure}

Next, we compare the average decision objective (revenue in this example) of each approach in Figure \ref{fig:revenue} over 100 test features. We then evaluate the decision using the ground truth demand corresponding to it. Note that we can evaluate ground truth only because the data is synthetic. We observe that our proposed robust approach outperforms the other methods by more than 30\% at larger $\epsilon$. At $\epsilon=0$, the method is not rboust (essentially by definition) and we observe similar performance as for the baselines.

The Gaussian Process method indeed improves in performance as we increase its noise parameter $\alpha$. However, it does not reach the accuracy of the other methods. The reward-learning method performs similarly to the baseline which learns demand.

\begin{figure}[t]
    \centering
    \includegraphics[scale=0.75]{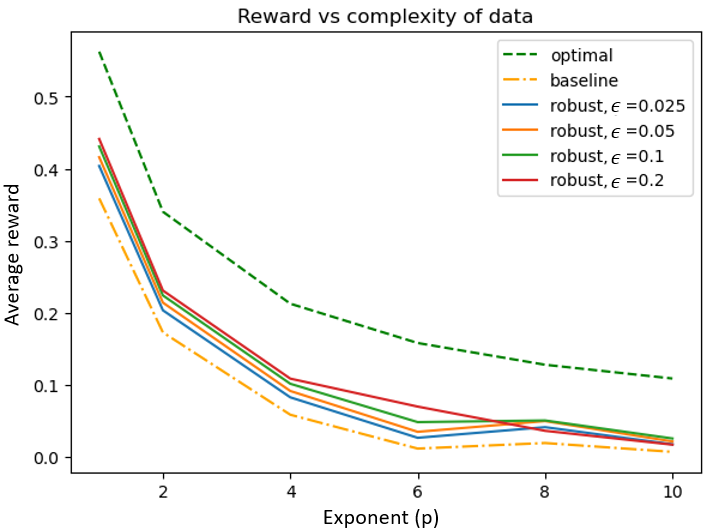}
    \caption{Varying demand-price complexity.}
    \label{fig:epsilon_v_exponent}
\end{figure}

\newpara{Effect of complexity $p$.} We now consider varying $p$, increasing the complexity between price and demand. Recall this parameter defined the demand function $\bar{z}_k = \lambda \cdot (B^T\x)^2_k + (1 - v_k)^p$ for item $k$. We compare the robust end-to-end method with a two-stage approach and the optimal method that knows the true distribution. We see the robust method outperforms the two-stage at all values of $p$. 
We also investigate the effect of varying $\epsilon$ as $p$ changes. See Figure \ref{fig:epsilon_v_exponent}. Notice that the most robust solution with $\epsilon=0.2$, performs best on data with lower complexity ($p=1,2,4,6)$ but worse at higher complexity ($p=8,10$) where the other lower values of $\epsilon$ perform better.  This experiment also exemplifies the method's stability, that we do not need extensive tuning to achieve good results. 


\subsection{Assortment}
\label{sec:ass}

For the assortment optimization problem we are given a set of $K$ products, and we must decide the amount $\v = (v_1, \dots, v_K)$ of each product to stock. The demand of a product type depends on its own stock as well as the stock of every other item. We also have a capacity constraint on the total amount of stock that can be stored: $\sum_{k=1}^K v_k \leq T$ for a fixed capacity $T$. Given a decision $\v$ and demand realization $\z = (z_1, \dots, z_K)$ for each product, the cost is given by 
\begin{equation}
  c(\v, \z) = \sum_{k=1}^K \max\{ z_k - v_k, 0 \} + b \cdot v_k. 
\end{equation}
There is a cost of $b \cdot v_k$ only to procure the items.
Moreover, we also consider a backorder cost for demand that is unmet by the inventory we chose to place, resulting in an additional cost of $\max\{ z_k - v_k, 0 \}$. If more inventory is placed than demand, there is no extra cost, all customers are satisfied. We assume the demand of an item is a function of the stock of all other items. Each item $k$ has a base demand $\alpha^*_k$ and some perturbation based on the other items $z_k = \alpha^*_k + \sum_{j\not=k}\beta^*_k \cdot v_k + \delta$ where $\delta$ is random normal noise. 
This captures complementary or supplementary item pair effects. Compared to the pricing problem, we have (1) higher dimension and hence interaction between the decisions that affect demand, and (2) a non-linear objective function which means that predicting the mean is not optimal (see our discussion after Eq. \eqref{def:task-loss}).


\newpara{Data generation.} We consider a more realistic generation of historical actions.  We assume the following: the store-owner tries to make the same inventory decision every week, however there is some probability that he will order a smaller quantity of a specific product. This skews the data we observe, and severely limits our observations compared to having uniform samples of all possible inventory choices. Robustness here is crucial to ensure that we do not pick decisions in regions that we have little to no data. 


\newpara{Methods.} We compare the performance of a linear model trained to learn the uncertainty, independent of the objective function $c$. This is the so-called \emph{predict-then-optimize} approach or a two-stage approach. The \emph{mean predictor} is an unrealistic model which exactly predicts the mean. The two-stage method should converge to this given enough data.  We compare with the linear model learned when optimizing jointly using the mixed-integer formulation \eqref{eq:solve-exact}. The formulation applied to this assortment problem can be found in Appendix \eqref{appendix:assortment}. We also compare using the sampling approach from \eqref{sec:algorithms1}. In addition, we compare against a pure reward-learning ML method, learning revenue directly as a function of $\v$. Finally, we also use a non-parameteric approach from \cite{bertsimas2022data} which uses a $K$-nearest neighbors (KNN) method to predict demand.

\begin{figure}
\centering
\includegraphics[width=0.95\linewidth,valign=B]{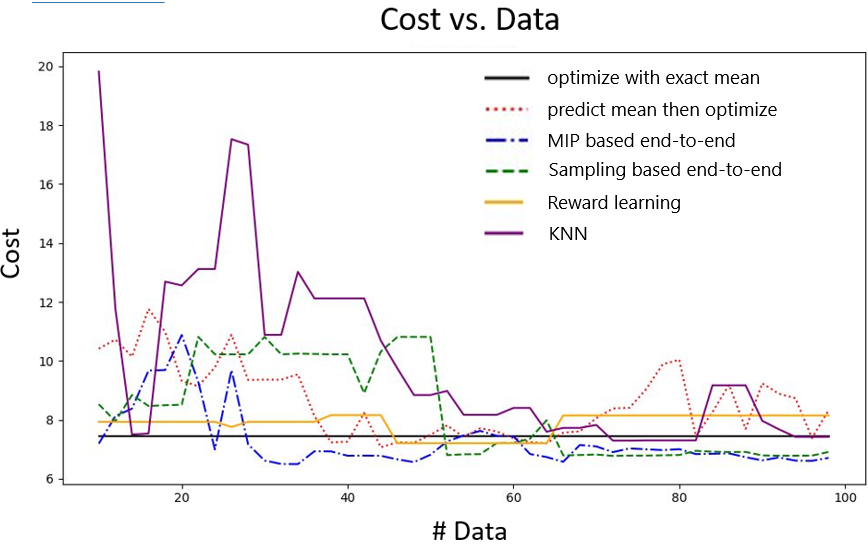} 
   \captionof{figure}{Assortment optimization: average cost across methods with training data.}
   \label{fig:assortment-data}
\end{figure}

\newpara{Results.} We see the results in Figure \ref{fig:assortment-data} for a 5-product assortment problem, as we vary the amount of data used for training. To evaluate the cost of a decision, we sample 1000 realizations of demand corresponding to the decision and report the average cost incurred. 
We observe the sampling-based approach requires more data, although is capable of reaching the same performance as the exact formulation. The main issue with two-stage approaches is the mismatch in objective --- they predict the mean of the demand distribution, but this is generally the incorrect statistic to predict. For example, if we want to learn some $f(\v)$ to predict demand, we wish that $c(\v, f(\v))$ is close to $\mathbb{E}_{\z \sim Z|\v} c(\v,\z)$. Since $c(\v,\z)$ is not linear in $\z$, we find that $\mathbb{E}_\z c(\v,\z) \not= c(\v, \mathbb{E}[\z])$. Indeed, for the single item case one can show that the optimal predictor $f(w)$ is the $((1-b)/b)^{th}$ quantile of the demand distribution. This is also why the \emph{mean predictor} is not optimal. The end-to-end approach learns a model $f$ which aligns $c(\v, f(\v))$ with $\mathbb{E}_{z} c(\v,\z)$. On the other hand, a reward-learning method performs worse since one need to learn a more complex map --- a nonlinear function of the demand function, whereas our approach only needs to learn a demand function. 

\newpara{Comparing with robust approaches.} We now analyze the performance of the robust approach from section \ref{section:robust}. We again compare against a \emph{two-stage} approach, which predicts demand by training a model that minimizes mean-squared error in demand prediction. We compare against a cost-learning approach, which instead learns cost $c(\v, \z)$ directly. This is the same methodology as the reward-learner in the pricing experiment. We also compare against the robust method we proposed, but solving the inner problem using the regularization technique as in \eqref{eq:reg}. We denote this as \emph{Robust-Regularization}. We also use our cutting-plane method and solve the inner problem using our proposed Algorithm \ref{alg:inner}. We denote this as \emph{Robust End-to-End}.

\newpara{Effect of robustness.} Now we analyze the decision cost of each model and we vary the robustness parameters as well. For our proposed approach, we vary $\epsilon$, the size of the uncertainty set. And for the \emph{Robust-Regularization} method we vary the robustness parameter $\lambda$ in \eqref{eq:reg}. First, we observe the difference in the cost-learning and 2-stage methods. For fairness, we use the same base model for all approaches. We indeed see that the 2-stage method performs better since the mapping from inventory to demand is simpler than inventory to reward (the objective value of $c$). Moreover, we see our proposed method of solving the inner problem (the blue curve) performs better than the regularization method (the purple curve). We train both methods using the same base model, the same number of epochs and hyperparameters (only varying $\epsilon, \lambda$). 

\begin{figure}
    \centering
    \includegraphics[width=0.8\linewidth]{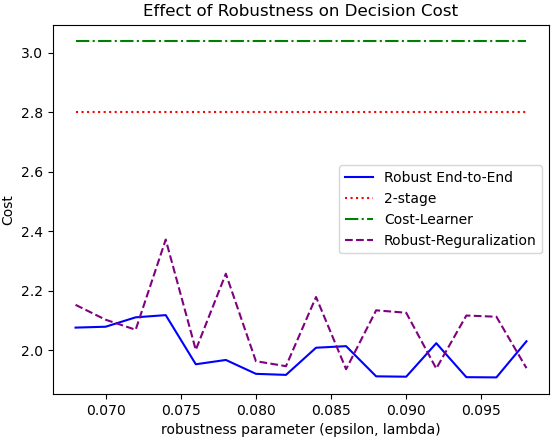}
    \caption{Decision cost on assortment problem as the robustness parameter increases, $\epsilon$ for Algorithm \ref{alg:inner}, and $\lambda$ in \eqref{eq:reg}.} 
    \label{fig:enter-label}
\end{figure}

\subsection{Electricity scheduling: information-gathering}
\label{sec:electricity}

\begin{figure}[b]
    \centering
    \includegraphics[scale=0.4]{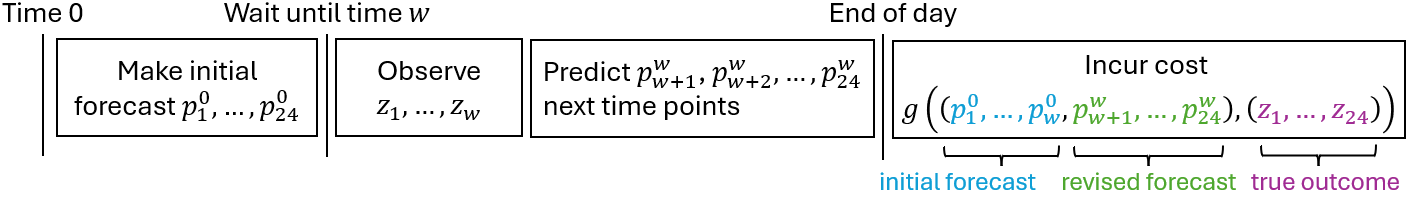}
    \caption{Electricity scheduling: sequence of actions. \vspace{-1em}}
    \label{fig:electricity-actions}
\end{figure}


We now consider the information-gathering setting introduced in section \ref{sec:info}. We consider an electricity generation scheduling problem using data from PJM, an electricity routing company coordinating the movement of electricity throughout 13 states. The goal is to make a generation schedule and decide on the amount of electricity to generate hour per hour, over the next 24 hours.
We consider the problem in two stages. First, we make an initial forecast for the 24 hours dependent only on feature information for that day. Then we decide on a time $w$ to update the schedule. Up to time $w$, we use the initial forecast to generate electricity, then given the new observations of true demand up to time $w$ we regenerate this forecast and generation schedule for the rest of the day. See figure \ref{fig:electricity-actions} for an illustration of the sequence of events. There is now a balancing act in deciding what hour $w$ to change the schedule. If we wait longer, we gain a better estimate of future demand, however we also use a worse forecast up to the waiting time $w$. Finally, we define the objective function. The operator incurs a unit cost $\gamma_e$ for excess generation and a cost $\gamma_s$ for shortages. The cost of generating $v_1, \dots, v_{24}$ while true demand is $z_1, \dots, z_{24}$ is given by
$
    c(\mathbf{v}, \z) = \sum_{i=1}^{24} \gamma_s \max\{ z_i - v_i, 0\} + \gamma_e \max\{ v_i - z_i, 0 \}.$

\begin{figure}[b]
    \centering
    \includegraphics[scale=0.4]{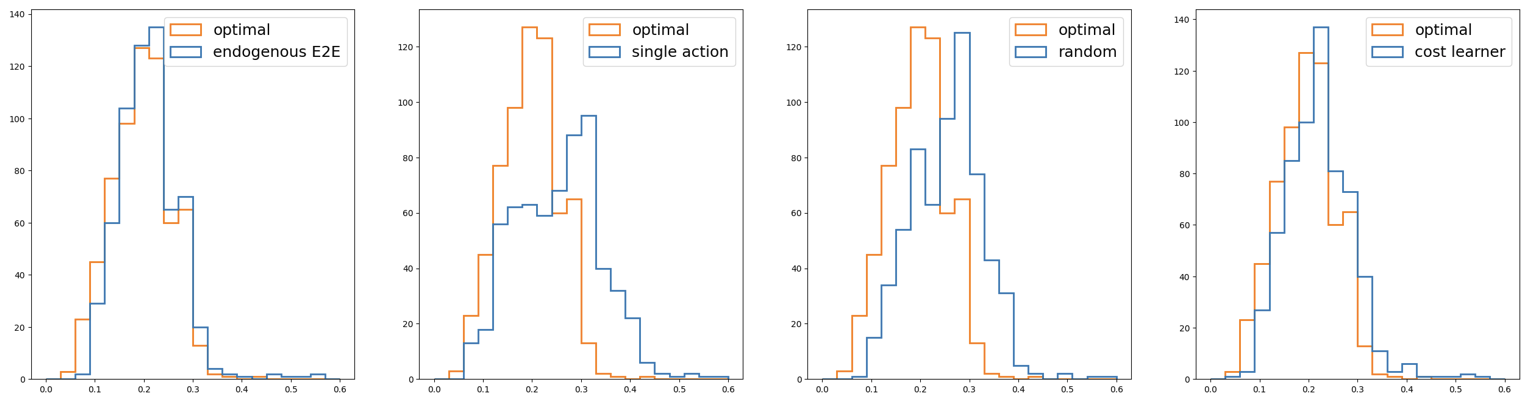}
    \caption{Cost distribution of each method. From left to right: endogenous E2E, single action, random, cost-learner.}
    \label{fig:cost-dist}
\end{figure}

\newpara{Methods:} Full details can be found in Appendix \ref{app:info}.   There are three components for each model: (1) how to make the initial forecast, (2) which time $w$ to choose, (3) how to update the forecast given $w$. 
We first introduce two baselines which always choose $w = 0$, so they never observe any of the day's demand, and only use their initial forecast. (1) We consider a predict-\emph{then}-optimize approach, where we learn a demand function independent of the decision-making step. We refer to this as ``Predict then optimize.'' (2) We learn an end-to-end model which aims to directly minimize decision cost $c$ as in \eqref{eq:exo-obj}. We refer to this as ``Vanilla E2E'' (vanilla end-to-end).

We then introduce baselines that choose $w$ in different ways, including our proposed approach. Each of these methods use the same model for the initial forecast, and for making the updated forecast (components (1) and (3) above). We choose the vanilla end-to-end method for this initial forecast since it performs significantly better than the 2-stage method. We only vary the method to decide $w$. The goal is to highlight the differences in objective cost resulting from various  methods of choosing $w$. 
Here we train a model $p$ to predict future demand given observations $z_1, \dots, z_w$, as well as features $\x$. 
Note that here we observe all variables up to time $w$, which is  different than in section \ref{sec:info} where we observe a single variable.  Up to time $w$, the baseline decisions are made by the vanilla end-to-end approach. After time $w$, the schedule is made according to $p$, based on \emph{true} demand up to time $w$. 

We have three baselines for methods on choosing $w$. (1) choosing a (uniformly) random action $w$. We refer to this as ``Random.'' 
(2) We fix a single $w$ for all data points (choosing this best $w$ from training data). We refer to this as ``Single action.'' Finally, (3) the optimal in-hindsight decision $w$ which may change for every out-of-sample data point $\x$. We refer to this as ``Optimal.'' 
We will denote our proposed method to decide $w$ as ``Endogenous E2E'' (endogenous end-to-end). 
This entails solving eq. \eqref{opt:endo1} by gradient descent for $f$ and choosing decision $w$ by solving eq. \eqref{eq:opt-info}. As a final baseline, we also compare against a more traditional approach: for each decision $w$ and features $\x$, predict directly the loss/cost of this decision. This does not take into account the structure of the problem and simply minimizes mean-squared error between predicted cost and observed cost of each action on the training data. We refer to this as ``Cost Learner.'' 

\newpara{Results:}  In table \ref{tab:exp-info}, we report the average difference between decision cost of our method and the other methods for each datapoint. Our approach is 7.5\% better than the cost-learning method and less than 8\% worse than optimal on average. We also measure the median cost of each method, as well as the percentage of test datapoints on which our approach performs better than every other approach. For example, our method only outperforms the predict-then-optimize method 66\% of the time, indicating this method does well on some data, but on also performs significantly more poorly on others, likely where it proposes a shortage (not knowing that a shortage is significantly worse than excess, since mean-squared error loss is unaware of this).  In addition, we also plot the cost distribution of each method on the test data in Figure \ref{fig:cost-dist} alongside the optimal cost distribution. We observe that our proposed method most closely aligns with the optimal cost everywhere. Knowing the additional structure of the problem, our approach can better learn it. 

\section{Conclusion}

This paper introduces an end-to-end framework for contextual stochastic optimization problems with decision-dependent uncertainty as well as for a class of two-stage information-gathering problems. In addition, we also propose a robust optimization approach to provide decisions that protect against more worst-case scenarios. We construct uncertainty sets over models whose task-based loss is bounded by some threshold, which we can think of as our budget of uncertainty allowed. 

We validate the approaches on a variety of problems, such as pricing, assortment optimization, and an information-gathering problem to decide how long to wait before rescheduling in an electricity generation task using real-world data. We compare against a variety of benchmarks as well, two-stage methods and robust approaches extended from offline learning and offline reinforcement learning. We consistently see improvement in performance in expected cost and robustness to uncertainty.

\begin{table}[t]
\centering
    \begin{tabular}[b]{l c c c}
        Method & \makecell{Average \\ difference} & \makecell{Median \\ cost} & \makecell{\% Endo. E2E \\ Wins} \\ \toprule
        {Predict then  optimize} & 710\% & 0.588 & 66\% \\
        Vanilla E2E & 55\% & 0.339 & 90\% \\
        Cost Learner & 7.5\% & 0.220 & 71\% \\
        \textbf{Endogenous E2E} & 0\% & $\mathbf{0.204}$ & 100\% \\ 
        Random & 21\% & 0.264 & 88\%  \\ 
        Single action & 20\% & 0.261 & 81\% \\ \midrule 
        Optimal & -7.9\% & 0.187 & NA
    \end{tabular}   
\caption{Electricity scheduling: cost comparison across methods.}
\label{tab:exp-info}
\end{table}


%
%
%

\newpage
\begin{APPENDICES}
\section{Experiment details}

Here we give more details on the experiments as well as the exact formulations used for these problems.

\subsection{Assortment}
\label{appendix:assortment}

For the assortment optimization problem we are given a set of $K$ products, and we must decide the amount $\v = (v_1, \dots, v_K)$ of each product to stock, given unknown demand $\z = (z_1, \dots, z_K)$. We are given data $(\v^1, \z^1), \dots, (\v^N, \z^N)$.
\begin{equation}
    c(\v, \z) = \sum_{k=1}^K \max\{ z_k - v_k, 0 \} + c \cdot v_k.
\end{equation}
Predictions of demand for a given decision is modeled by 
\begin{equation}
    f_{\alpha,\beta}(\v)_k = \alpha_k + \sum_{j\not=k} \beta_{k,j} v_j
\end{equation}
and we aim to solve the empirical prediction problem 
\begin{align}
    \min_{\alpha, \beta} \sum_{n=1}^N &\ (c(\v^n, f_{\alpha, \beta}(\v^n)) - c(\v^n, \z^n))^2    \\
    \min_{\alpha, \beta} \sum_{n=1}^N &\ \paren{\paren{\sum_{k=1}^K \max\{ \alpha_k + \sum_{j\not=k} \beta_{k,j}v^n_j - v^n_k, 0\} + c \cdot v^n_k } - c(\v^n, \z^n)}^2.
\end{align}
To solve this problem, we introduce continuous variables $w_{n,k}$ and binary variables $y_{n,k}$ to enforce that $y_{n,k} = 1$ whenever $f_{\alpha,\beta}(v^n) - v^n_k \geq 0$ (and 0 otherwise) and that always $w_{n,k} = f_{\alpha,\beta}(\v^n) - \v^n_k$. We do so using the following constraints:
\begin{align}
    w_{n,k} \geq&\ f_{\alpha,\beta}(\v^n) - v^n_k \\
    w_{n,k} \geq&\ 0 \\
    w_{n,k} \leq&\ (f_{\alpha,\beta}(\v^n) - v^n_k) + M(1-y_{n,k})\\
    w_{n,k} \leq&\ (0) + My_{n,k}.
\end{align}
Here, $M$ needs to be a constant large enough so that $|f_{\alpha, \beta}(\v^n)| \leq M$ for all $\v^n$. In practice, we can choose this heuristically by looking at the scale of the magnitude of the data on $\z^n$. With these constraints, the objective can be rewritten as 
\begin{equation}
    \min_{\alpha, \beta} \sum_{n=1}^N \paren{\paren{\sum_{k=1}^K w_{n,k}} - c(\v^n, \z^n)}^2.
\end{equation}

\subsection{Information-gathering: electricity scheduling}
\label{app:info}

We use a similar set-up for the problem as in \cite{donti2017task}. This paper only considered the pure single-stage end-to-end task, and not the two-stage problem we are considering here with information gathering. Nevertheless, we use the same data, model architecture, and similar problem parameters which we describe here.

\emph{The set-up:} We are asked to decide/plan on the amount of electricity to generate each hour for the next 24 hours. We denote these decisions by $v_1, \dots, v_{24}$. Given a demand realization of $z_1, \dots, z_{24}$, the cost of the decision is 
\begin{equation}
    c(\v, \z) = \sum_{i=1}^{24} \gamma_s \max\{ \z_i - v_i, 0\} + \gamma_e \max\{ v_i - \z_i, 0 \}.
\end{equation}
Each day, we are given contextual information $\x^n$ and demand observation $\mathbf{d}^n$. As features, we use the past day's electrical load as well as temperature, and the temperature forecast for the current day. In addition, we use non-linear functions of the temperature, one-hot-encodings of holidays and weekends, and yearly sinusoidal features. Like the paper \cite{donti2017task}, we use a 2-layer feed-forward network, both hidden layers having width 200, and an additional residual connection from the input to the output. This linear layer is initialized by first solving a linear regression problem to predict demand (this is done independently of the objective $c(\cdot, \cdot)$). 

This is the original single-stage end-to-end problem. In our setting we also consider the possibility to update the schedule as the day progresses. Each day, we must decide ahead of time a particular hour to regenerate the schedule. We denote this decision by $w \in \{ 0, 1, \dots, 24 \}$. For the first $w$ hours of the day, we use the original decisions made the day before. After observing $z_1, \dots, z_w$, we then make a new forecast conditioned on these observations, and make a new decision based off of this for the remaining hours $w+1, \dots, 24$.

\emph{Methods:}  The base prediction model for each method is the same across all approaches. We use the two-layer network described in the above paragraphs. Let $\mathcal{F}$ denote this architecture and the set of models/weights using this architecture. Any model $f \in \mathcal{F}$ takes as input features $\x$ and outputs a vector in $\R^{24}_{\geq 0}$ for each hour of the next day. 

\begin{enumerate}
    \item \textbf{Two-Stage}: This is a simple regression model which predicts demand as a function of features $\x$. That is, we train
    \begin{equation}
        f_{\text{2-stage}} = \arg\min_{f \in \mathcal{F}} \sum_{n=1}^N (f(\x^n) - \z^n)^2.
    \end{equation}
    Then, for an out-of-sample $\x$, we simply set decisions $v_1, \dots, v_n$ according to $ f_{\text{2-stage}}(\x) $.
    \item \textbf{Vanilla End-to-End}: Here the objective is to directly minimize the cost of the decisions we take. So, instead of minimizing mean-squared error, the objective is to minimize cost:
    \begin{equation}
        f_{\text{end-to-end}} = \arg\min_{f \in \mathcal{F}} \sum_{n=1}^N c(f(\x^n), \z^n).
    \end{equation}
    Again, we take decisions decisions $v_1, \dots, v_n$ according to $ f_{\text{2-stage}}(\x) $. 

    \item \textbf{Learning $\bm{p}$}: Before describing the remaining methods, we first focus on the model $p(\x, z_1, \dots, z_w)$. Here, we have chosen to wait until hour $w$, then we observe $z_1, \dots, z_w$. Given these, would like to make a new forecast for the remaining hours of the day. This forecast is given by $p(\x, z_1, \dots, z_w)$. Since the input length of $\hat{p}$ varies with $w$, we take the full vector $\z$ as input (which always has fixed length 24) and mask the time points from $w+1$ to $24$. Moreover, $\hat{p}$ still outputs the full 24 time points, but the loss function will only be evaluated on time points $w+1$ to $24$.

    We first use the initial vanilla end-to-end predictions as a baseline. Let these be $f_{\text{end-to-end}}(\x)$. Then, the model $p$ will predict a perturbation to this baseline, dependent only on the new observations $z_1, \dots, z_w$. Specifically, let 
    \begin{equation}
        p(\x, z_1, \dots, z_w) = f_{\text{end-to-end}}(\x) + \hat{p}(z_1, \dots, z_w) 
    \end{equation}
    and we wish to learn this $\hat{p}(z_1, \dots, z_w)$. For our experiments, we let $\hat{p}$ be a single hidden layer network of width 200 with relu activation. 
    
    At each batch of training, the loss function is given by 
    \begin{equation}
        \frac{1}{24-w} c_{w+1, \dots, 24}(f_{\text{end-to-end}}(\x) + \hat{p}(z_1, \dots, z_w), \z)
    \end{equation}
    where at each batch, we randomly choose a different $w$ and $c_{w+1, \dots, 24}(\hat{\z}, \z)$ denotes the objective function evaluated only on time points starting from $w+1$ (since there is no need to evaluate on the first $w$ time points). The leading term $1/(24-w)$ is meant to take the average cost per hour.

    \item \textbf{Random action}: We now consider a model which randomly chooses $w$ for every datapoint. There are two scheduling decisions: (1) the schedule up to time $w$ and (2) the schedule after time $w$. The initial schedule is determined by the vanilla end-to-end method, then given observations $z_1, \dots, z_w$ at time $w$, we use the model $p(\x, z_1, \dots, z_w)$ to decide the rest of the schedule. Let $\hat{\z}$ denote this combined schedule. The cost of the action is then determined by $c(\hat{\z}, \z)$.

    \item \textbf{Single action}: Here we choose a single fixed across $w$ across all datapoints. We choose this action to be the one that results in the lowest cost on training data. We determine the cost of an action $w$ in the same way as for the \emph{random action} method above.

    \item \textbf{Optimal action}: For every datapoint, we compute the cost of every possible action $w = 0, \dots, 24$ and choose the best one. 

    \item \textbf{Endogenous end-to-end}: We now train a model $f^e(\x, w)$ with the goal that using this as a point forecast when making decision $w$. Here, the superscript $e$ denotes endogenous, to separate from previous functions like $f_{\text{end-to-end}}, f_{\text{2-stage}}$. This is done as follows.
    \begin{enumerate}
        \item We make a point forecast $f^{e}(w, \x)$.
        \item Given point forecast, we predict $p(\x, f^{e}_{1}(\x, w), \dots, f^{e}_{w}(\x, w))$ for time points after time $w$.
        \item For ground truth, we would observe $z_1, \dots, z_w$. We would use this instead of $f^{e}_{1}(\x, w), \dots, f^{e}_w(\x, w)$ when making a forecast for the second stage. That is, we would predict $p(\x, z_1, \dots, z_w)$ instead.  
        \item The second-stage schedule is given by $p(\x, f^{e}_{1}(\x, w), \dots, f^{e}_{w}(\x, w))$ while the schedule given ground truth observations is given by $p(\x, z_1, \dots, z_w)$. The loss function is then the squared difference between the \emph{cost} of these decisions. Specifically, this is 
    \begin{equation}
        \paren{c_{w+1, \dots, 24}(p(\x, f^{e}_{1}(\x, w), \dots, f^{e}_{w}(\x, w)), f^{e}(\x, w)) - c(p(\x, z_1, \dots, z_w), \z)}^2
    \end{equation}
    where again $c_{w+1, \dots, 24}(\hat{\z}, \z)$ only evaluates the loss starting at time point $w+1$, omitting the first $w$.
    \item Again, at every batch, we randomly choose some fixed $w$ for all datapoints in the batch.
    \end{enumerate}
    Now, given learned $f$ and $p$, we must make a decision according to \eqref{eq:opt-info}. When applied to the electricity scheduling problem, we solve 
    \begin{equation}
    \begin{split}
        \min_{w \in \{0,\dots,24\}} & c_{1,\dots,w}(f_{\text{end-to-end}}(\x), f^{e}(w, \x))  + \\ & c_{w+1,\dots,24}(p(w, f^{e}_1(w, \x), \dots, f^{e}_w(w, \x)), f^{e}(w, \x))         
    \end{split}
    \end{equation}
    where the first term evaluates predicted cost of the initial schedule before time $w$ and the second term evaluates predicted cost on the schedule after time $w$.

    \item \textbf{Cost learner}: Here the model learns the cost of each action $w$ given $\x$. That is, a model $f_{\text{cost}}(w)$ wich predicts
    \begin{equation}
        f_{\text{cost}}(\x, w) \approx c_{1,\dots,w}(f_{\text{end-to-end}}(\x), \z) + c_{w+1, \dots, 24}(p(w, z_1, \dots, z_w), \z)
    \end{equation}
    the cost of using the vanilla end-to-end schedule up to time $w$, and then the $p$ schedule after time $w$. To make decisions, we choose $w$ which minimizes $f_{\text{cost}}(\x, w)$.
 
    \item \textbf{Evaluate on test data:} Finally, for any decision $w$, we evaluate as follows on test data. We use the base vanilla end-to-end schedule for the first $w$ time points. We then observe the ground truth $z_1, \dots, z_w$ and make predictions $p(\x, z_1, \dots, z_w)$ to make the rest of the schedule. Formally, the cost is 
    \begin{equation}
        c_{1,\dots,w}(f_{\text{end-to-end}}(\x), \z) + c_{w+1, \dots, 24}(p(w, z_1, \dots, z_w), \z),
    \end{equation}
    exactly the same as the cost-learner's target above.
\end{enumerate}

\section{Proofs}
\label{sec:proofs}

\begin{proof}{Proof of Proposition \ref{prop:exist}}
    Consider two values $\z^1$ and $\z^2$ so that 
    \begin{equation}
c(\v, {\z}^1) \leq \mathbb{E}_{\v \in Z|\v, \x} [c(\v, \z)] \leq 
c(\v, {\z}^2).
    \end{equation}
Since $c$ is a continuous function with respect to $\z$, there must exist a convex combination of $\z^1, \z^2$, say $\hat{\z}$ so that 
\begin{equation}
c(\v, \hat{\z}) = \mathbb{E}_{\z \in Z|\v, \x} [c(\v, \z)].
\end{equation}

$\qed$
\end{proof}

\begin{proof}{Proof of Theorem \ref{thm:generalization}}
 The results of \cite{bartlett2002rademacher} can be applied directly to the composite cost function $c(\hat{\z}) = (c(\v, \hat{\z}) - \y)^2$  where for simplicity we use $\y$ to replace the constant $c(\v,\z)$. The loss of a model $f \in \mathcal{F}$ is given by the $c \circ f = c(f(\v, \x))$. Theorem 8 of \cite{bartlett2002rademacher} gives us 
 \begin{equation}
 \label{eq:rad-thm}
 l(f) \leq \hat{l}(f) +  \mathscr{R}_N(c \circ \mathcal{F}) + \paren{\frac{8\log 2/\delta}{N}}^{1/2}.
 \end{equation}
 Next, using the vector contraction inequality from \cite{bartlett2002rademacher}, we can further bound the Rademacher complexity by 
 \begin{equation}
    \label{eq:r-bound}
     \mathscr{R}_N(c \circ \mathcal{F}) \leq \sqrt{2} \lambda \mathscr{R}_N(\mathcal{F})
 \end{equation}
 where the cost function $g(\hat{\z}) = (c(\v, \hat{\z}) - \y)^2$ is $\lambda$-Lipschitz with respect to $\hat{\z}$. It remains to bound $\lambda$. Any continuously differentiable function over a compact domain is Lipschitz continuous with Lipschitz constant equal to the maximum magnitude of the derivative over that domain. In our case, $c(\cdot, \cdot) \in [0,1]$. 

 We can further decompose $g(\hat{z})$ into $g_1 \circ g_2$ where $g_1(\z') = (\z')^2$ and $g_2(\hat{\z}) = c(\v, \hat{\z}) - \y$. The Lipschitz constant of $g(\hat{\z})$ is then bounded by the product of the Lispchitz constants of $g_1$ and $g_2$. By assumption (in theorem \ref{thm:generalization}), $c$ is $L$-Lipschitz and hence so is $g_2(\hat{\z})$. Moreover, $\z' = c_1(\hat{\z}) \in [-1,1]$ since both $c(\cdot, \cdot)$ and $\y$ are in $[0,1]$. Next, $c_1$ is 2-Lipschitz since its gradient is $2\z'$ and its greatest magnitude is $|2\z'| \leq 2$ over $\z' \in [-1,1]$. Therefore, $\lambda \leq 2 \cdot L$. This combined with \eqref{eq:r-bound} and \eqref{eq:rad-thm} proves our theorem. 

 $\qed$
 
\end{proof}

\begin{proof}{Proof of Theorem \ref{theorem:uncertainty}}
    Let $\mathcal{N}_{\gamma}$ be a $\gamma$-covering of $\mathcal{F}$. The proof consists of two main steps. (1) to relate the probability that $f^* \in \mathcal{U}_{\epsilon}$ to the probability that $f^* \in \mathcal{N}_\gamma$. Then, (2) to bound the latter probability directly. We treat the random draw of the dataset as the random variable. We denote this random variable by $\mathcal{H}$.

    For ease of notation, let $\hat{\mathcal{E}}(f)$ denote the empirical error of $f$. Specifically, 
    \begin{equation}
        \hat{\mathcal{E}}(f) = \frac{1}{N} \sum_{n=1}^N \paren{ c(\v_n, f(\v_n, \x_n)) - c(\v_n, \z_n) }^2
    \end{equation}
so that we can succintly describe $\mathcal{U}_\epsilon = \{ f \in \mathcal{F} : \hat{\mathcal{E}}(f) \leq \hat{\mathcal{E}}(\hat{f}_{\text{opt}}) + \epsilon \}$ where $ \hat{\mathcal{E}}(\hat{f}_{\text{opt}}) $ is the minimizer over $\mathcal{F}$ of the empirical error. We wish to provide an upper bound on the probability that $\hat{\mathcal{E}}(f^*) > \hat{\mathcal{E}}(\hat{f}_{\text{opt}}) + \epsilon$. To do so, we prove an upper bound for  similar bound for all $\tilde{f} \in \mathcal{U}_\epsilon$:
\begin{equation}
    \mathbb{P}_{\mathcal{H}}\paren{\hat{\mathcal{E}}(f^*) > \hat{\mathcal{E}}(\hat{f}_{\text{opt}}) + \epsilon} \leq \mathbb{P}_{\mathcal{H}}\paren{\bigcup_{\tilde{f} \in \mathcal{U}_\epsilon} \{ \hat{\mathcal{E}}(f^*) > \hat{\mathcal{E}}(\tilde{f}) + \epsilon \} } 
\end{equation}
Note that the event $\bigcup_{\tilde{f} \in \mathcal{U}_\epsilon} \{ \hat{\mathcal{E}}(f^*) > \hat{\mathcal{E}}(\tilde{f}) + \epsilon \}$ is equivalent to the event that $\hat{\mathcal{E}}(f^*) > \min_{\tilde{f} \in \mathcal{U}_\epsilon} \hat{\mathcal{E}}(\tilde{f}) + \epsilon$. Moreover, for any $\tilde{f} \in \mathcal{U}_\epsilon$, there exists some $h_{\tilde{f}} \in \mathcal{N}_\gamma$ for which $\norm{\tilde{f} - h_{\tilde{f}}}_{d} \leq \gamma $ (where recall $d$ is the metric defined on $\mathcal{F}$). Then,
\begin{equation}
    |\hat{\mathcal{E}}(\tilde{f}) - \hat{\mathcal{E}}(h_{\tilde{f}})| \leq 3M\gamma
\end{equation}
Now, we can write 
\begin{align}
\min_{\tilde{f} \in \mathcal{U}_\epsilon} \hat{\mathcal{E}}(\tilde{f}) + \epsilon =&\ \min_{\tilde{f} \in \mathcal{U}_\epsilon} \hat{\mathcal{E}}(\tilde{f}) - \hat{\mathcal{E}}(h_{\tilde{f}}) + \hat{\mathcal{E}}(h_{\tilde{f}}) + \epsilon \\ 
\geq&\ \hat{\mathcal{E}}(h_{\tilde{f}}) + \epsilon - 3M\gamma \\
\geq&\ \min_{h \in \mathcal{N}_\gamma} \hat{\mathcal{E}}(h) + \epsilon - 3M\gamma 
\end{align}
Therefore, whenever the event $\bigcup_{\tilde{f} \in \mathcal{U}_\epsilon} \{ \hat{\mathcal{E}}(f^*) > \hat{\mathcal{E}}(\tilde{f}) + \epsilon \}$ occurs, so must $\bigcup_{h \in \mathcal{N}_\gamma} \{ \hat{\mathcal{E}}(f^*) > \hat{\mathcal{E}}(h) + \epsilon - 3M\gamma  \}$. It now suffices to provide an upper bound over the discrete set $\mathcal{N}_\gamma$:
\begin{equation}
\label{eq1-1}
\mathbb{P}_{\mathcal{H}}\paren{\bigcup_{\tilde{f} \in \mathcal{U}_\epsilon} \{ \hat{\mathcal{E}}(f^*) > \hat{\mathcal{E}}(\tilde{f}) + \epsilon \} }     \leq \mathbb{P}_{\mathcal{H}}\paren{ \bigcup_{h \in \mathcal{N}_\gamma} \{ \hat{\mathcal{E}}(f^*) > \hat{\mathcal{E}}(h) + \epsilon - 3M\gamma  \} } 
\end{equation}

We now proceed to prove an upper bound for any single even occuring for any fixed $h \in \mathcal{N}_\gamma$ then we take a union bound to prove the claim. To do so, we make use of concentration bounds for martingale difference sequences. We define the following random variables. Let, 
\begin{align}
    U^i(f) =&\ \paren{c(\v_i, f(\v_i, \x_i)) - c(\v_i, \z_i)}^2 \\
    V^i(f) =&\ U^i(f) - U^i(f^*) \\
    T^i(f) =&\ \mathbb{E}V^i(f) - V^i(f)
\end{align}
Notice that the sequence $(T^i(f))_{i=1}^n$ is a martingale difference sequence. We make use of the following concentration lemma:
\begin{lemma}
    Let $(\x_i)_{i\leq n}$ be a real-valued martingale difference sequence. If $|\x_i|\leq 1$ almost surely, then for any $\eta \in [0,1]$, with probability at least $1 - \delta$, for all $t \leq n$, 
    \begin{equation}
        \sum_{i=1}^t \x_i \leq \eta \sum_{i=1}^t \mathbb{E}[\x_i^2] + \frac{\log \delta^-1}{\eta}
    \end{equation}
\end{lemma}
Therefore, choosing $\eta = 1/8$ we have for all $t = 1, \dots, n$, with probability $1 - \delta$,
\begin{equation}
    \sum_{i=1}^t T^i(f) \leq \frac{1}{8}  \sum_{i=1}^t \mathbb{E}[T^i(f)^2] + 8\log \frac{1}{\delta}
\end{equation}
To simplify the term $ \mathbb{E}[T^i(f)^2] $, we bound this as 
\begin{align}
     \mathbb{E}[T^i(f)^2] =&\ \mathbb{E}[(\mathbb{E}V^i(f) - V^i(f))^2] \\ 
     =&\ \mathbb{E}[V^i(f)^2] - \mathbb{E}V^i(f))^2  \\
     \leq&\ \mathbb{E}[V^i(f)^2] \\ 
     \leq&\ \mathbb{E}\left[ (U^i(f) - U^i(f^*))^2 \right] \\ 
     \leq&\ \mathbb{E}\left[ \paren{  (c(\v_i, f(\v_i,\x_i)) - c(\v_i,\z_i))^2 - (c(\v_i, f^*(\v_i,\x_i)) - c(\v_i,\z_i))^2 }^2 \right] \\ 
     \leq&\ \mathbb{E}\left[ \paren{  c(\v_i, f(\v_i,\x_i))^2 - c(\v_i, f^*(\v_i,\x_i))^2 - 2c(\v_i,\z_i)( c(\v_i, f(\v_i,\x_i)) - c(\v_i, f^*(\v_i,\x_i))) }^2 \right] \\ 
     \leq&\  \mathbb{E}\left[ \paren{ [c(\v_i, f(\v_i,\x_i)) - c(\v_i, f^*(\v_i,\x_i))]\cdot[c(\v_i, f(\v_i,\x_i)) + c(\v_i, f^*(\v_i,\x_i)) - 2c(\v_i, \z_i)]   }^2 \right] 
\end{align}
Now since $c(\v_i, f(\v_i,\x_i)), c(\v_i, f^*(\v_i,\x_i)), c(\v_i, \z_i) \leq 1$ by assumption, we can bound the right term to be in the range $[-2,2]$. Therefore,
\begin{equation}
    \mathbb{E}[T^i(f)^2] \leq 4\mathbb{E} \left[ \paren{c(\v_i, f(\v_i,\x_i)) - c(\v_i, f^*(\v_i,\x_i))}^2 \right]
\end{equation}
We now show that the right hand side is equal to $4\mathbb{E}[V^i(f)]$. Indeed, 
\begin{align}
    \mathbb{E}[V^i(f)] =&\ \mathbb{E}\left[ \paren{c(\v_i, f(\v_i, \x_i)) - c(\v_i, \z_i)}^2 - \paren{c(\v_i, f^*(\v_i, \x_i)) - c(\v_i, \z_i)}^2 \right] \\ 
    =&\ \mathbb{E}\left[ c(\v_i, f(\v_i,\x_i))^2 - c(\v_i, f^*(\v_i,\x_i))^2 - 2c(\v_i,\z_i)(c(\v_i, f(\v_i,\x_i)) - c(\v_i, f^*(\v_i,\x_i))) \right]
\end{align}
Recall by definition that $\mathbb{E}[c(\v_i,\z_i)] = c(\v_i, f^*(\v_i,\z_i))$. Moreover, notice that $f(\v_i,\x_i)$ and $f^*(\v_i,\x_i)$ are independent of $\z_i$ since $f,f^*$ only depend on the history up to $i-1$. Therefore, we can simplify the above to 
\begin{align}
    \mathbb{E}[V^i(f)] =&\ \mathbb{E}\left[  c(\v_i, f(\v_i,\x_i))^2 - 2 c(\v_i, f(\v_i,\x_i)) c(\v_i, f^*(\v_i,\x_i)) +  c(\v_i, f^*(\v_i,\x_i))^2 \right] \\
    =&\ \mathbb{E} \left[ \paren{c(\v_i, f(\v_i,\x_i)) - c(\v_i, f^*(\v_i,\x_i))}^2 \right]    
\end{align}
Therefore, with probability at least $1-\delta$,
\begin{align}
    \sum_{i=1}^t T^i(f) \leq&\ \frac{1}{8}  \sum_{i=1}^t \mathbb{E}[T^i(f)^2] + 8\log \frac{1}{\delta} \\     
    \sum_{i=1}^t \mathbb{E}V^i(f) - V^i(f) \leq&\ \frac{1}{2}  \sum_{i=1}^t \mathbb{E}V^i(f) + 8 \log\frac{1}{\delta} \\ 
    \frac{1}{2} \sum_{i=1}^t \mathbb{E}V^i(f) \leq&\ \sum_{i=1}^t V^i(f) + 8 \log\frac{1}{\delta}
\end{align}
Note that $\mathbb{E}V^i(f) = \mathbb{E}U^i(f) - \mathbb{E}U^i(f^*) \geq 0$ since $f^*$ is optimal. So, with probability at least $1 - \delta$,
\begin{align}
    0 \leq&\ \sum_{i=1}^t V^i(f) + 8 \log\frac{1}{\delta} \\
    \sum_{i=1}^t U^i(f^*) \leq&\ \sum_{i=1}^t U^i(f) + 8 \log\frac{1}{\delta} \\ 
\end{align}
and 
\begin{equation}
    \mathbb{P}\paren{ \hat{\mathcal{E}}(f^*) > \hat{\mathcal{E}}(f) + \frac{8}{t} \log \frac{1}{\delta} } \leq \delta
\end{equation}
Applying union bound, 
\begin{equation}
    \mathbb{P}\paren{ \bigcup_{h \in \mathcal{N}_\gamma} \left\{ \hat{\mathcal{E}}(f^*) > \hat{\mathcal{E}}(h) + \frac{8}{t} \log \frac{|\mathcal{N}_\gamma|}{\delta} \right\} } \leq \delta
\end{equation}
Therefore, setting $\epsilon = \frac{8}{t} \log \frac{|\mathcal{N}_\gamma|}{\delta} + 3M\gamma $ in equation \eqref{eq1-1} proves the theorem. We now prove the corollary as follows.

When $\mathcal{F}$ is the space of $d$-dimensional linear functions with bounded norm (that is, $\norm{f} \leq \alpha$ for all $f \in \mathcal{F}$), we have that $|\mathcal{N}_\gamma| \leq \paren{1 + 1/\gamma}^d$. One can choose $\gamma$ in any way, independent of any implementation considerations. When choosing $\gamma = 1/\log N$, the result follows.

$\qed$

\end{proof} 

\end{APPENDICES}

\ACKNOWLEDGMENT{}


\bibliographystyle{informs2014} 
\bibliography{refs} 

\begin{thebibliography}{28}
\providecommand{\natexlab}[1]{#1}
\providecommand{\url}[1]{\texttt{#1}}
\providecommand{\urlprefix}{URL }

\bibitem[{Agrawal et~al.(2019{\natexlab{a}})Agrawal, Amos, Barratt, Boyd,
  Diamond, \protect\BIBand{} Kolter}]{agrawal2019differentiable}
Agrawal A, Amos B, Barratt S, Boyd S, Diamond S, Kolter JZ (2019{\natexlab{a}})
  Differentiable convex optimization layers. \emph{Advances in neural
  information processing systems} 32.

\bibitem[{Agrawal et~al.(2019{\natexlab{b}})Agrawal, Avadhanula, Goyal,
  \protect\BIBand{} Zeevi}]{agrawal2019mnl}
Agrawal S, Avadhanula V, Goyal V, Zeevi A (2019{\natexlab{b}}) Mnl-bandit: A
  dynamic learning approach to assortment selection. \emph{Operations Research}
  67(5):1453--1485.

\bibitem[{Akchen \protect\BIBand{} Misic(2021)}]{akchen2021assortment}
Akchen YC, Misic VV (2021) Assortment optimization under the decision forest
  model. \emph{arXiv preprint arXiv:2103.14067} .

\bibitem[{Amos \protect\BIBand{} Kolter(2017)}]{amos2017optnet}
Amos B, Kolter JZ (2017) Optnet: Differentiable optimization as a layer in
  neural networks. \emph{International conference on machine learning},
  136--145 (PMLR).

\bibitem[{Arumugam \protect\BIBand{} Van~Roy(2021)}]{arumugam2021value}
Arumugam D, Van~Roy B (2021) The value of information when deciding what to
  learn. \emph{Advances in Neural Information Processing Systems}
  34:9816--9827.

\bibitem[{Bartlett \protect\BIBand{} Mendelson(2002)}]{bartlett2002rademacher}
Bartlett PL, Mendelson S (2002) Rademacher and gaussian complexities: Risk
  bounds and structural results. \emph{Journal of Machine Learning Research}
  3(Nov):463--482.

\bibitem[{Berthet et~al.(2020)Berthet, Blondel, Teboul, Cuturi, Vert,
  \protect\BIBand{} Bach}]{berthet2020learning}
Berthet Q, Blondel M, Teboul O, Cuturi M, Vert JP, Bach F (2020) Learning with
  differentiable pertubed optimizers. \emph{Advances in neural information
  processing systems} 33:9508--9519.

\bibitem[{Bertsimas \protect\BIBand{} Koduri(2022)}]{bertsimas2022data}
Bertsimas D, Koduri N (2022) Data-driven optimization: A reproducing kernel
  hilbert space approach. \emph{Operations Research} 70(1):454--471.

\bibitem[{Chu et~al.(2011)Chu, Li, Reyzin, \protect\BIBand{}
  Schapire}]{chu2011contextual}
Chu W, Li L, Reyzin L, Schapire R (2011) Contextual bandits with linear payoff
  functions. \emph{Proceedings of the Fourteenth International Conference on
  Artificial Intelligence and Statistics}, 208--214 (JMLR Workshop and
  Conference Proceedings).

\bibitem[{Cohen \protect\BIBand{} Perakis(2019)}]{cohen2019optimizing}
Cohen MC, Perakis G (2019) Optimizing promotions for multiple items in
  supermarkets. \emph{Channel strategies and marketing mix in a connected
  world}, 71--97 (Springer).

\bibitem[{Cristian et~al.(2023)Cristian, Harsha, Perakis, Quanz,
  \protect\BIBand{} Spantidakis}]{cristian2023end}
Cristian R, Harsha P, Perakis G, Quanz BL, Spantidakis I (2023) End-to-end
  learning for optimization via constraint-enforcing approximators.
  \emph{Proceedings of the AAAI Conference on Artificial Intelligence},
  volume~37, 7253--7260.

\bibitem[{Donti et~al.(2017)Donti, Amos, \protect\BIBand{}
  Kolter}]{donti2017task}
Donti P, Amos B, Kolter JZ (2017) Task-based end-to-end model learning in
  stochastic optimization. \emph{Advances in neural information processing
  systems} 30.

\bibitem[{Elmachtoub \protect\BIBand{} Grigas(2022)}]{elmachtoub2022smart}
Elmachtoub AN, Grigas P (2022) Smart “predict, then optimize”.
  \emph{Management Science} 68(1):9--26.

\bibitem[{Ettl et~al.(2020)Ettl, Harsha, Papush, \protect\BIBand{}
  Perakis}]{ettl2020data}
Ettl M, Harsha P, Papush A, Perakis G (2020) A data-driven approach to
  personalized bundle pricing and recommendation. \emph{Manufacturing \&
  Service Operations Management} 22(3):461--480.

\bibitem[{Ferber et~al.(2020)Ferber, Wilder, Dilkina, \protect\BIBand{}
  Tambe}]{ferber2020mipaal}
Ferber A, Wilder B, Dilkina B, Tambe M (2020) Mipaal: Mixed integer program as
  a layer. \emph{Proceedings of the AAAI Conference on Artificial
  Intelligence}, volume~34, 1504--1511.

\bibitem[{Foster \protect\BIBand{} Rakhlin(2020)}]{foster2020beyond}
Foster D, Rakhlin A (2020) Beyond ucb: Optimal and efficient contextual bandits
  with regression oracles. \emph{International Conference on Machine Learning},
  3199--3210 (PMLR).

\bibitem[{Foster et~al.(2021)Foster, Kakade, Qian, \protect\BIBand{}
  Rakhlin}]{foster2021statistical}
Foster DJ, Kakade SM, Qian J, Rakhlin A (2021) The statistical complexity of
  interactive decision making. \emph{arXiv preprint arXiv:2112.13487} .

\bibitem[{Fujimoto et~al.(2019)Fujimoto, Meger, \protect\BIBand{}
  Precup}]{fujimoto2019off}
Fujimoto S, Meger D, Precup D (2019) Off-policy deep reinforcement learning
  without exploration. \emph{International conference on machine learning},
  2052--2062 (PMLR).

\bibitem[{Howard(1966)}]{howard1966information}
Howard RA (1966) Information value theory. \emph{IEEE Transactions on systems
  science and cybernetics} 2(1):22--26.

\bibitem[{Kash et~al.(2023)Kash, Key, \protect\BIBand{}
  Zoumpoulis}]{kash2023optimal}
Kash IA, Key PB, Zoumpoulis SI (2023) Optimal pricing and introduction timing
  of technology upgrades in subscription-based services. \emph{Operations
  Research} 71(2):665--687.

\bibitem[{Kelley(1960)}]{kelley1960cutting}
Kelley JE Jr (1960) The cutting-plane method for solving convex programs.
  \emph{Journal of the society for Industrial and Applied Mathematics}
  8(4):703--712.

\bibitem[{Kim et~al.(2024)Kim, Iyengar, \protect\BIBand{}
  Zeevi}]{kim2024doubly}
Kim W, Iyengar G, Zeevi A (2024) A doubly robust approach to sparse
  reinforcement learning. \emph{International Conference on Artificial
  Intelligence and Statistics}, 2305--2313 (PMLR).

\bibitem[{Kumar et~al.(2019)Kumar, Fu, Soh, Tucker, \protect\BIBand{}
  Levine}]{kumar2019stabilizing}
Kumar A, Fu J, Soh M, Tucker G, Levine S (2019) Stabilizing off-policy
  q-learning via bootstrapping error reduction. \emph{Advances in Neural
  Information Processing Systems} 32.

\bibitem[{Kumar et~al.(2020)Kumar, Zhou, Tucker, \protect\BIBand{}
  Levine}]{kumar2020conservative}
Kumar A, Zhou A, Tucker G, Levine S (2020) Conservative q-learning for offline
  reinforcement learning. \emph{Advances in Neural Information Processing
  Systems} 33:1179--1191.

\bibitem[{Mandi et~al.(2020)Mandi, Stuckey, Guns et~al.}]{mandi2020smart}
Mandi J, Stuckey PJ, Guns T, et~al. (2020) Smart predict-and-optimize for hard
  combinatorial optimization problems. \emph{Proceedings of the AAAI Conference
  on Artificial Intelligence}, volume~34, 1603--1610.

\bibitem[{Sadana et~al.(2023)Sadana, Chenreddy, Delage, Forel, Frejinger,
  \protect\BIBand{} Vidal}]{sadana2023survey}
Sadana U, Chenreddy A, Delage E, Forel A, Frejinger E, Vidal T (2023) A survey
  of contextual optimization methods for decision making under uncertainty.
  \emph{arXiv preprint arXiv:2306.10374} .

\bibitem[{Vlastelica et~al.(2019)Vlastelica, Paulus, Musil, Martius,
  \protect\BIBand{} Rol{\'\i}nek}]{vlastelica2019differentiation}
Vlastelica M, Paulus A, Musil V, Martius G, Rol{\'\i}nek M (2019)
  Differentiation of blackbox combinatorial solvers. \emph{arXiv preprint
  arXiv:1912.02175} .

\bibitem[{Xie et~al.(2021)Xie, Cheng, Jiang, Mineiro, \protect\BIBand{}
  Agarwal}]{xie2021bellman}
Xie T, Cheng CA, Jiang N, Mineiro P, Agarwal A (2021) Bellman-consistent
  pessimism for offline reinforcement learning. \emph{Advances in neural
  information processing systems} 34:6683--6694.

\end{thebibliography}





\end{document}